\documentclass[final,5p,times,twocolumn]{elsarticle}

\usepackage{amssymb}
\usepackage{amsmath}
\usepackage{subcaption} 
\usepackage{multirow}
\usepackage{hyperref}
\usepackage{threeparttable} 
\usepackage{tablefootnote} 
\usepackage[table,xcdraw]{xcolor} 
\usepackage{booktabs}    
\usepackage{multirow}    
\usepackage{array}       
\usepackage{graphicx}    
\usepackage{caption}     
\usepackage{float}       
\usepackage{setspace}    
\usepackage{geometry}    
\usepackage{xcolor}      
\usepackage{placeins}  

\begin{document}

\begin{frontmatter}



\title{MK-SGN: A Spiking Graph Convolutional Network with Multimodal Fusion and Knowledge Distillation for Skeleton-based Action Recognition} 
{\color{black}
\author[1]{
    Naichuan Zheng
}
\ead{2022110134zhengnaichuan@bupt.edu.cn}
\author[1]{Hailun Xia\corref{cor1}}
\ead{xiahailun@bupt.edu.cn}
\cortext[cor1]{Corresponding author.}
\author[1]{ 
    Zeyu Liang
}
\ead{lzy_sfading@bupt.edu.cn}
\author[1]{ 
    Yuchen Du
}
\ead{duyuchen@bupt.edu.cn}
\affiliation[1]{
    organization={Beijing Laboratory of Advanced Information Networks, Beijing Key Laboratory of Network System Architecture and Convergence, School of Information and Communication Engineering, Beijing University of Posts and Telecommunications},
    addressline={}, 
    city={Beijing},
    postcode={100876}, 
    state={},
    country={China}
}
}
\begin{abstract}
In recent years, multimodal Graph Convolutional Networks (GCNs) have achieved remarkable performance in skeleton-based action recognition. The reliance on high-energy-consuming continuous floating-point operations inherent in GCN-based methods poses significant challenge for deployment in energy-constrained, battery-powered edge devices. To address  the limitation, MK-SGN, a Spiking Graph Convolutional Network with Multimodal Fusion and Knowledge Distillation, is proposed to leverage the energy efficiency of Spiking Neural Networks (SNNs) for skeleton-based action recognition for the first time. By integrating the energy-saving properties of SNNs with the graph representation capabilities of GCNs, MK-SGN achieves significant reduction in energy consumption while maintaining competitive recognition accuracy.
Firstly, we formulate a Spiking Multimodal Fusion (SMF) module to fuse the multimodal skeleton spike-form feature effectively.   Secondly, we propose a Self-Attention Spiking Graph Convolution (SA-SGC) module and a Spiking Temporal Convolution (STC) module to capture spatial relationships and temporal dynamics of spike-form features, respectively. Finally, we propose an integrated knowledge distillation strategy to transfer information from the multimodal GCN to the SGN, incorporating inner-layer feature and soft-label distillation to enhance the SGN's performance.
MK-SGN exhibits substantial advantages, surpassing state-of-the-art GCN frameworks in energy efficiency and outperforming state-of-the-art SNN frameworks in recognition accuracy. The proposed method remarkably reduces energy consumption, with 98\% less consumed than conventional GCN-based approaches. In general, this research establishes a strong baseline for developing high-performance, energy-efficient SNN-based models for skeleton-based action recognition.
\end{abstract}




\end{frontmatter}




\section{Introduction}
\label{sec1}

Human action recognition is a critical task with wide-ranging applications in human-computer interaction, identity verification, and emergency detection~\cite{zhang2019}. Initial approaches predominantly relied on RGB-based methods that analyzed visual information extracted from video frames\cite{khaire2022}. While effective in certain contexts, such methods encountered significant challenges in complex environments characterized by cluttered backgrounds, inadequate illumination, and occlusions, limiting their robustness and generalization capabilities.
\begin{figure}[!h]
  \centering
  \includegraphics[width=0.9\linewidth]{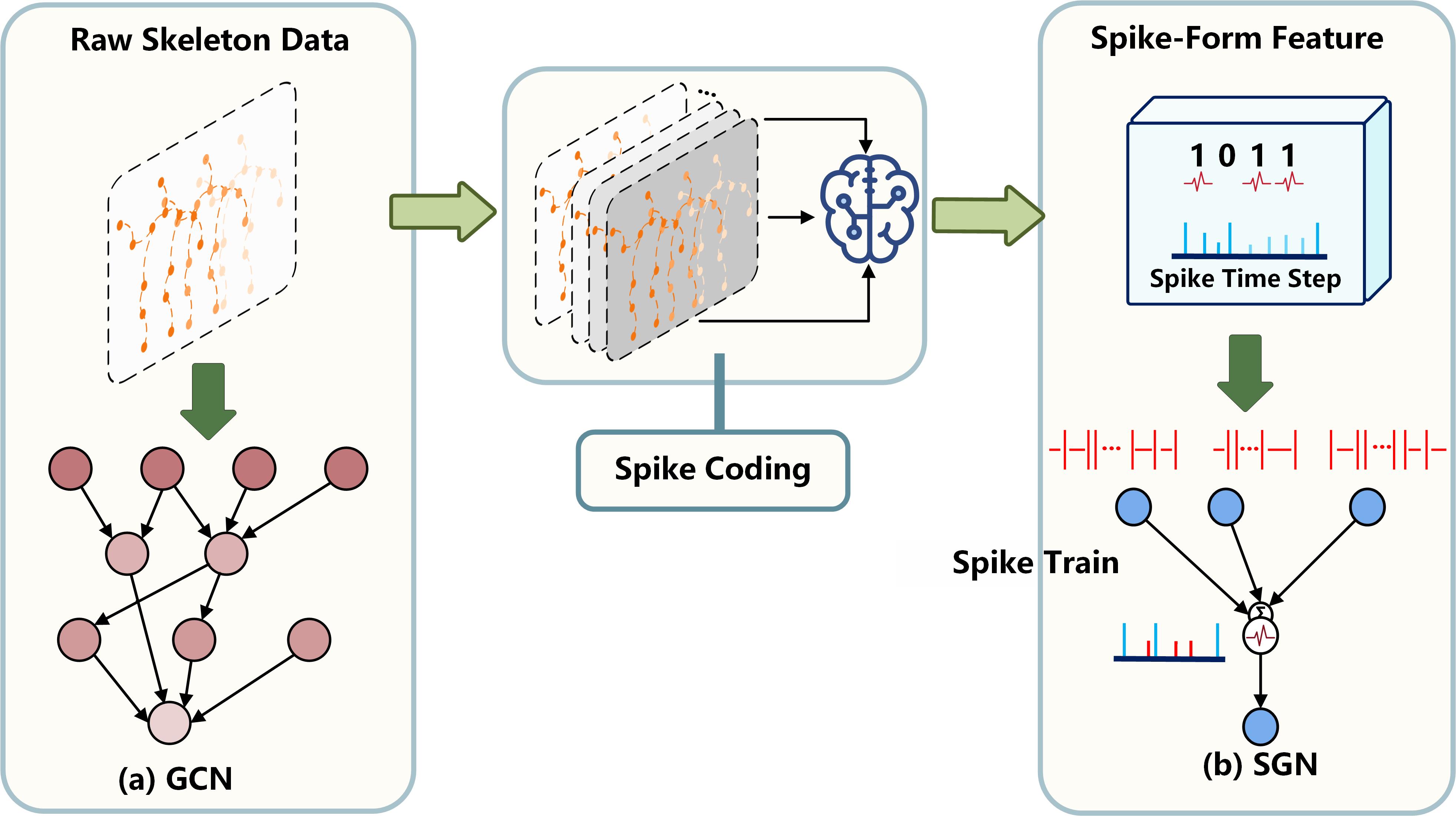}
\caption{\textcolor{black}{Transformation from skeleton to spikes. The former GCN performs feature extraction
and information propagation on nodes of the graph structure through graph convolutional layers, while the latter SGN employs spike encoding to convert features into discrete
spike-form, conducting feature extraction and information
propagation in the time dimension of the spike-form feature.}}
\label{fig:trans}
\end{figure}
Research has shifted towards skeleton-based action recognition to address limitations in these earlier approaches, which utilize 3D skeleton data obtained from depth sensors and pose estimation\cite{ren2024}. By isolating human action dynamics from environmental disturbances, skeleton-based recognition enhances robustness and reliability in action recognition tasks. Early methods employed Convolutional Neural Networks (CNNs) \cite{zhu2019, ding2017, li2017}, Recurrent Neural Networks (RNNs) \cite{du2015, du2016}, and Long Short-Term Memory networks (LSTMs) \cite{liu2017, zhang2018} to model spatio-temporal features. However, limitations in representing the intrinsic structure of skeleton data prompted the development of Graph Convolutional Networks (GCNs), which represent a significant advancement in the field \cite{ahmad2021}.

GCNs effectively model the spatial and temporal relationships within skeleton data, with the introduction of ST-GCN representing a key milestone \cite{yan2018}. The architecture employs spatial-temporal graph convolution to process skeleton data, laying the foundation for more advanced frameworks. 2S-AGCN improves performance by integrating a two-stream structure (Joint and Bone Streams) alongside adaptive graph convolution, enabling dynamic learning of skeleton topologies and refinement of temporal features \cite{shi2019}. Subsequent advancements aim to balance efficiency and accuracy. Shift-GCN utilizes a channel-wise shift operation to simulate spatio-temporal information flow with lightweight computational overhead\cite{cheng2020}. CTR-GCN introduces channel-level topology optimization and attention mechanisms to enhance feature representation \cite{chen2021}.
Building on these, Info-GCN leverages information bottleneck theory for representation learning and extends the skeleton modality to six streams using the k-hop approach \cite{chi2022}. The development culminated in HULK, a multimodal large-scale model designed to unify diverse modalities within a robust framework. By harnessing its extensive capacity, HULK achieves state-of-the-art(SOTA) performance in precision and multimodal integration \cite{wang2023}.

However, such methods rely on deep network architectures typical of artificial neural networks (ANNs), which depend heavily on computationally expensive floating-point operations. Such reliance results in considerable computational energy demands. For example, lightweight models like Shift-GCN and CTR-GCN consume 46 mJ and 36.25 mJ per action sample, respectively, while 2S-AGCN requires up to 182.87 mJ per sample (single-sample energy consumption calculated based on Equation~\ref{equation:power}). Additionally, the pursuit of better performance necessitates the integration of multimodal data, which requires separate training and testing for each modality, followed by ensemble fusion. These characteristics of ANN-based methods lead to significant computational complexity and energy consumption, posing substantial challenges for deployment on battery-powered edge devices and other energy-constrained environments.

To address the high energy consumption and deployment challenges associated with ANN-based methods, Spiking Neural Networks (SNNs) provide a promising alternative, renowned for their energy efficiency \cite{tavanaei2019}. Inspired by biomimetics, SNNs utilize event-driven mechanisms that enable exceptionally low energy consumption, making them particularly well-suited for resource-constrained environments. The energy-efficient design positions SNNs as a viable solution to the limitations of traditional ANN frameworks in practical applications \cite{yamazaki2022, ponulak2011, rathi2023}. SNNs have evolved significantly alongside advancements in ANNs, incorporating advanced architectures such as Spiking Graph Neural Networks \cite{yin2024, zhao2024, zhu2022}, Spiking Recurrent Neural Networks \cite{yin2021, yin2020, xing2020}, and Spiking Transformers \cite{zhou2022, yao2024a, yao2024b}. Research efforts have primarily focused on tasks such as RGB image classification \cite{niu2023} and event-driven recognition \cite{zhang2022, cordone2022, yao2021}, achieving notable success in reducing energy consumption while maintaining high accuracy.

Building on advancements in SNNs and GCNs, we propose a novel Spiking Graph Network (SGN) framework that, for the first time, integrates the energy efficiency of SNNs with the graph representation capabilities of GCNs for skeleton-based action recognition. Our work establishes the first framework leveraging SNNs for skeleton-based action recognition, significantly advancing this task's potential in energy-constrained environments.

The exploration of SGNs for skeleton action recognition is guided by several scientific challenges that prompt deep reflection:

\begin{enumerate}
    \item \textbf{Transformation of Skeleton Data into Spike-based Representations:} 
    Effectively transforming skeleton data, characterized by complex graphical structures, into spike-based representations compatible with the computational principles of SNNs remains a crucial challenge.

    \item \textbf{Multimodal Feature Fusion within SNNs:} 
    Facilitating the fusion of multimodal features in SNNs requires advanced methods to ensure consistency, coherence, and resilience in feature representation.
    
    \item \textbf{Compatibility with Vanilla GCNs:} 
    Adapting the vanilla GCN framework to spike-based representations while ensuring robust performance in action recognition poses a significant research challenge.

    \item \textbf{Optimization of Energy Efficiency and Accuracy:} 
    Balancing energy consumption and recognition accuracy to meet the requirements of energy-constrained environments remains an essential goal.
\end{enumerate}
To address these issues, we first propose a Skeleton Spiking Coding (SSC) module to convert skeleton data into event spikes and a Spike-based Multimodal Fusion (SMF) module based on mutual information to compute the correlation between multimodal data for early fusion. The SMF module fully leverages the complementary information from multimodal skeleton data while avoiding the heavy training and testing burden associated with ensemble methods. Based on these components, we propose a novel SGN block integrating a Self-Attention-Spiking Graph Convolutions (SA-SGC) module and a Spiking Temporal Convolutions (STC) module. The SA-SGC captures the spatial dependencies of skeleton sequences represented in spiking form through a graph topology and enhances them via a spiking attention mechanism, while the STC extracts temporal dependencies to complement the spatial information represented by the SA-SGC. This integrated design provides a unified framework that leverages the strengths of SGN and vanilla GCNs for effective feature extraction and information propagation.
Additionally, a GCN-to-SGN knowledge distillation method is formulated, comprising inner-layer feature distillation and soft label distillation. High-precision GCN models guide the training of SGN through these complementary strategies. A Feature Translation Module (FTM) is formulated to convert inner-layer features from multimodal GCN into spike-form features, enabling effective inner-layer feature distillation. Soft label distillation transfers the output distribution knowledge from GCN to SGN, further enhancing model performance. The dual-distillation approach improves accuracy without increasing model complexity. Leveraging the maturity of GCN architectures and pre-trained weights, the process imposes no additional training burden.
Ultimately, we propose Spiking Graph Convolutional Network with Multimodal Fusion and Knowledge Distillation (MK-SGN), the first framework to leverage SNNs for skeleton-based action recognition. Extensive experiments validate the effectiveness of the proposed approach, demonstrating its ability to achieve energy-efficient and accurate performance while bridging the gap between SNNs and GCNs in this domain.

Overall, the main contributions of this paper can be summarized as follows:

(1) A novel MK-SGN framework is proposed, introducing skeleton-based action recognition into the domain of SNNs and establishing a foundation for future research in energy-efficient action recognition.

(2)An SMF module is formulated to fuse multimodal skeleton data in spike form based on mutual information, effectively integrating complementary information while reducing energy consumption and computational overhead. The novel SA-SGC and STC modules further enhance the representation of spatial relationships and temporal dynamics within the fused spike-form features, respectively.

(3)A knowledge distillation strategy is established to transfer knowledge from GCN to SGN, integrating inner-layer feature and soft-label distillation for heterogeneous networks. To enable effective inner-layer feature distillation, an FTM is employed to convert features from the intermediate layers of multimodal GCN into spike-form representations.

(4) Extensive experiments on three challenging datasets demonstrate the advantages of our work. The theoretical energy consumption of MK-SGN is \textbf{0.596 mJ per action sample}. In general, Compared to SOTA GCN-based methods, \textbf{with 98\% less energy consumed than conventional GCN-based approaches}. Moreover, we surpass the accuracy of SOTA SNN-based methods.
    
\section{Related Work}
\label{subsec1}
\subsection{Skeleton-based Action Recognition}
In recent years, deep learning has significantly advanced skeleton-based action recognition \cite{ren2024}. Early research primarily employed CNNs to extract spatial features, capturing joint positions and local spatial relationships \cite{zhu2019, ding2017, li2017}. RNNs, especially LSTMs, were then applied to model temporal dynamics, excelling at capturing long-term sequential dependencies in motion data \cite{du2015, du2016, zhang2018, liu2017spatio, zheng2019}. While these approaches successfully addressed spatial and temporal dimensions separately, they failed to fully exploit the inherent graph structure of skeleton data, limiting their capacity to represent complex joint dependencies and hierarchical relationships.

The introduction of ST-GCN \cite{yan2018} marked a turning point by representing skeletons as graphs and applying graph convolutions to jointly capture spatial and temporal patterns. However, its reliance on a fixed adjacency matrix restricted adaptability to dynamic joint interactions. To overcome this limitation, subsequent studies proposed learnable adjacency matrices \cite{xie2021, shi2020, lee2023}, which allow edges to be optimized for specific actions, though at the cost of higher model complexity and potential overfitting. Another line of work incorporates attention mechanisms \cite{liu2023, wu2024selfgcn, wu2024localglobal, tian2024}, which selectively emphasize informative joints and capture long-range dependencies, thereby enriching feature representations. Nevertheless, these methods often increase computational cost. Other adaptive strategies, such as multi-scale sampling or channel-mixing, further improve robustness and flexibility, but the balance among accuracy, efficiency, and generalizability remains an open challenge.

\textcolor{black}{
Beyond spatial modeling, researchers have also enhanced temporal modeling. MS-G3D \cite{liu2020} introduced multi-scale temporal modeling through dilated temporal windows and cross-spacetime graph convolutions, improving the capture of long-range temporal dependencies, while STA-GCN \cite{hang2022} used temporal adaptive graph convolutions to jointly model short- and long-term patterns.Together, these advances demonstrate the importance of refining both spatial and temporal receptive fields to improve overall recognition performance.}

\textcolor{black}{
More recently, multimodal integration has become a key driver of progress. Approaches that incorporate joint, bone, motion, and other dynamic cues enrich feature diversity. With the development of multimodal foundation models, Hulk \cite{wang2023} unified heterogeneous skeleton modalities into a comprehensive framework, achieving state-of-the-art accuracy. BlockGCN \cite{zhou2024} dynamically redefined topology awareness to better capture dependencies, offering accuracy comparable to Hulk while retaining lower computational cost. This contrast highlights the trade-off between large-scale multimodal integration and efficient graph-based modeling. The rise of large language models (LLMs) has also inspired novel directions. PURLS \cite{zhu2024} leveraged GPT-3 to enhance zero-shot action recognition by generating semantic action descriptions, facilitating transfer to unseen classes. Neuron \cite{chen2025} advanced this line with evolving micro-prototypes for tighter alignment between skeleton data and semantic context. Similarly, CrossGLG \cite{yan2024} applied a global–local–global LLM-guided strategy to one-shot recognition, enriching feature abstraction across multiple levels. Although not directly targeting action recognition, M-R²ET \cite{zhang2024} explored motion retargeting through decoupled skeleton and shape modeling, offering valuable insights into handling diverse skeleton structures and transfer learning.}

\textcolor{black}{
Despite these advances, most existing methods remain grounded in conventional ANNs, which rely heavily on floating-point operations and thus incur high computational and energy costs. This limits their scalability and deployment in resource-constrained scenarios. SNNs, known for their event-driven and energy-efficient computation, present a promising alternative. However, their potential in skeleton-based action recognition has not yet been explored, leaving a critical research gap that our work aims to address.}

\subsection{Spiking Neural Networks}
Unlike ANNs, SNNs transmit information through discrete spike sequences, a mechanism inspired by biological principles. This spike-driven mode closely mimics brain function, offering significant potential for energy efficiency \cite{tavanaei2019}. Early research in SNNs focused on various spiking neuron models designed to convert continuous inputs into spike sequences. Notable models include the Integrate-and-Fire (IF) neuron \cite{tal1997}, which simplifies the process by accumulating input signals until a threshold is reached. The Hodgkin-Huxley (HH) model \cite{hodgkin1952} provides a biophysically detailed representation of neuronal activity based on ion channel dynamics.
The Leaky Integrate-and-Fire (LIF) neuron is widely regarded for its simplicity and efficiency \cite{wu2018}. The dynamics and update of the membrane potential are described as follows:
\begin{equation}
\tau_m \frac{d V}{d t}=-\left(V-V_{\text {rest }}\right)+R \cdot I(t) ,
\end{equation}
\begin{equation}
V(t+\Delta t) = 
\begin{cases} 
V_{\text{reset}}, & \text{\hspace{-2em}if } V(t) \geq V_{\text{thresh}} \\
V(t) + \left( \frac{-\left(V(t) - V_{\text{rest}}\right) + I(t)}{\tau_m} \right) \Delta t, & \text{otherwise}
\end{cases},
\end{equation}
where \(V(t)\) represents the membrane potential at the current time point, where \(V_{\text{rest}}\) denotes the neuron's resting potential.  \(I(t)\) indicates the input current. The membrane time constant is given by \(\tau_m\), and  \(V_{\text{thresh}}\) is the threshold potential that triggers the firing of an action potential. After firing, the membrane potential is reset to \(V_{\text{reset}}\). The model's discrete update time step is \(\Delta t\). \\
\textcolor{black}{
\indent These advancements have led to the development of various spiking neural architectures, such as Spiking Graph Convolutional Networks  \cite{yin2024,zhao2024,zhu2022}, which integrate Spiking Neural Networks  into Graph Convolutional Networks to efficiently process dynamic graph data while reducing computational costs. In the domain of sequence processing, Spiking Transformers \cite{zhou2022,yao2024a,yao2024b} leverage spiking self-attention mechanisms to replace traditional matrix multiplication with sparse operations, improving energy efficiency in tasks like RGB image processing. Meanwhile, Spiking Recurrent Neural Networks\cite{yin2021,yin2020,xing2020} combine the temporal dynamics of recurrent networks with spiking neuron models, offering superior performance in sequential tasks such as speech and gesture recognition, while significantly reducing power consumption. Lastly, Spiking Convolutional Neural Networks \cite{cordone2021,zhang2021} adapt spiking neurons to convolutional layers, enabling efficient processing of spatial correlations in image-based tasks, providing an energy-efficient alternative for real-time applications on neuromorphic hardware.\\
\indent Studies in this area have demonstrated substantial energy savings without compromising accuracy\cite{{niu2023}}.
Other research has focused on event-driven recognition, leveraging SNNs for their suitability in processing sparse and asynchronous data \cite{vicente-sola2025}. Although SNNs have made significant progress in other areas, their application to skeleton-based action recognition remains underexplored. The event-driven nature of SNNs provides a promising approach to significantly improving energy efficiency, albeit with an inevitable trade-off in accuracy. Leveraging this potential could provide a viable solution to balancing energy efficiency and accuracy in action recognition tasks.}

\section{Proposed Method}
This section provides an overview of MK-SGN's comprehensive architecture, followed by a detailed explanation of its Skeleton Spiking Coding module and Spiking Multimodal Fusion Module. Additionally, the SGN Block and the Knowledge Distillation Method from GCN to SGN are thoroughly discussed.
\begin{figure*}[t]
  \centering
  \includegraphics[width=1\linewidth]{all.jpg}
  \caption{\textcolor{black}{The overview of the proposed MK-SGN architecture. The teacher network comprises a 10-layer GCN that processes four multimodal skeleton data separately, followed by Global Average Pooling (GAP) and Fully Connected (FC) layers to generate soft labels. The student network is a 6-layer SGN that incorporates the following key modules: (a) The Skeleton Spiking Coding (SSC) module, responsible for transforming raw skeleton data into spike-form feature to ensure compatibility with spiking neural computations. (b) The Spiking Self-Attention Spiking Graph Convolution (SA-SGC) module is formulated to model spatial dependencies and enhance feature extraction by integrating Spiking Graph Convolution and Spiking Self-Attention (SSA). (c) The Spiking Temporal Convolution (STC) module is proposed to capture temporal patterns and refine temporal dynamics within spike-form features. Additionally, the Spike-based Multimodal Fusion Module (SMF) optimizes cross-modal integration based on mutual information, and the Feature Transformation Module (FTM) maps inter-layer GCN features into spike-form features for effective alignment with the SGN. Knowledge distillation is achieved through the FTM and supervision from the soft labels produced by the teacher network.}}
\label{fig:MK-SGN}
\end{figure*}
\subsection{Overall Architecture}
Our network architecture, illustrated in Figure~\ref{fig:MK-SGN}, comprises a 10-layer multimodal teacher GCN modal and a 6-layer student SGN modal. 
The teacher GCN processes the four multimodal skeleton data separately. The processed features from each modality are then passed through the Global Average Pooling (GAP) and Fully Connected (FC) layers to generate predictions. These predictions are finally fused to produce the soft labels.
The student SGN network processes each modality through the SSC module, transforming skeleton data into spike-form representations. Spike-form features are then fused via the SMF module, which utilizes mutual information to optimize cross-modal integration. Each student network layer includes the SA-SGC and STC modules, which are formulated to capture and refine spatial and temporal features efficiently and effectively.
Our approach implements knowledge distillation through inter-layer features and soft label distillation. The inter-layer features from the teacher GCN modal are converted into spike-form representations via the FTM, which ensures compatibility with the spiking neural domain and facilitates the training of the student SGN modal. Soft labels produced by the teacher modal provide supplementary supervisory signals, optimizing the learning process. The subsequent sections provide a detailed account of the formulation of the modules.

\subsection{Spiking Coding Mechanism and Spiking Multimodal Fusion}
This chapter explores the conversion from continuous real-valued skeleton data into spiking-form representations. It then details the methods for fusing multimodal spike-form feature derived from multimodal skeleton data.
\begin{figure}[t]
    \centering
    \begin{subfigure}{0.25\textwidth}
        \centering
        \includegraphics[width=\textwidth]{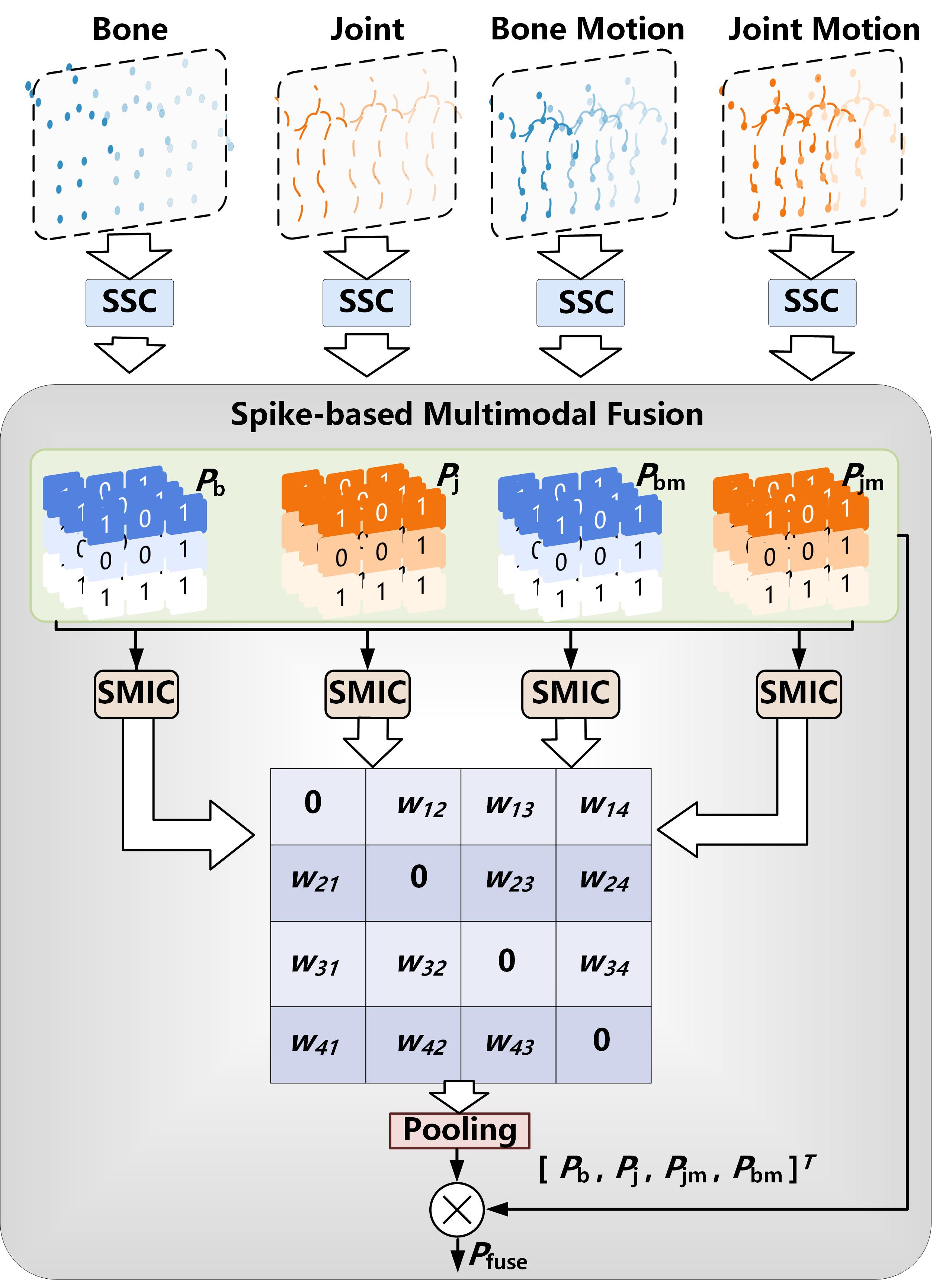}
        \caption{\textcolor{black}{Multimodal Skeleton Spiking Coding and Mutual Information Weight Matrix}}
        \label{fig:sub1}
    \end{subfigure}
    \hfill
    \begin{subfigure}{0.18\textwidth}
        \centering
        \includegraphics[width=\textwidth]{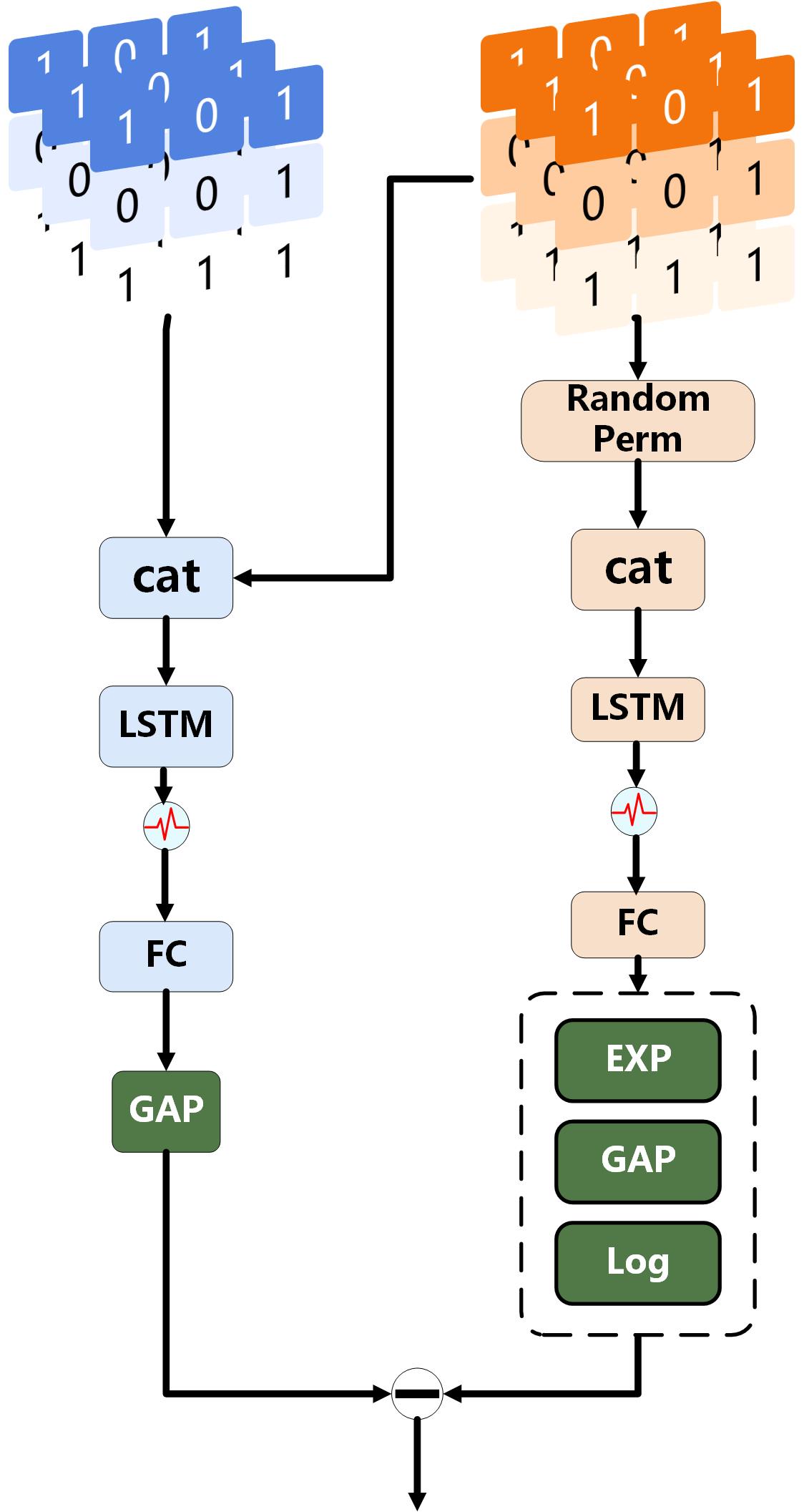}
        \caption{\textcolor{black}{The detail structure of the Spike-form Mutual Information Calculation (SMIC) Module}}
        \label{fig:sub2}
    \end{subfigure}
    \caption{\textcolor{black}{Multimodal Skeleton Spiking Coding and Mutual Information Weight Matrix Calculation Process}}
    \label{fig:main}
\end{figure}
\subsubsection{Skeleton Spiking Coding Module}
For a given 3D skeleton sequence data \(X_{\mathrm{i}} \in \mathbb{R}^{C \times T \times V}\), the SSC Module first transforms it into an image-like representation. Then, it expands it into a form suitable for spiking temporal representation illustrated in Figure~\ref{fig:MK-SGN}(a). The process is described as follows:
\begin{equation}
P_0 = \operatorname{Expand}(X_\text{i}) \in \mathbb{R}^{S\times C \times V \times T},   
\end{equation}
\begin{equation}
P_{\text{i}} = \operatorname{SN}(\operatorname{BN}(\operatorname{Conv2d}(P_0))) \in \mathbb{R}^{S \times D \times V \times T}.
\end{equation}
In the initial step, the skeleton sequence data is transformed into an image-like representation to facilitate the structured feature extraction of spatial and temporal characteristics in subsequent stages. Subsequently, the $\operatorname{Expand}(\cdot)$ operation adds a spike time step dimension  \(S\)  to the data, reorganizing the temporal structure to align with the requirements of spiking representation. The expanded representation \(P_0\) incorporates the additional spike time step dimension \(S\) while preserving the original temporal dimension \(T\), the graph structure \(V\), and the channel size \(C\).

Following expansion, the representation \(P_0\) is processed through a 2D convolutional layer (\(\operatorname{Conv2d}\)) with a kernel size of 3, followed by Batch Normalization (\(\operatorname{BN}\)) and a Spiking Neuron (\(\operatorname{SN}\)) layer. The pipeline refines the temporal-spatial features within the spiking domain, yielding a spiking feature representation \(P\) with dimensions \(S \times D \times V \times T\). The spike time step dimension \(S\) reflects the temporal structure of spiking activity, while the dimension \(T\) retains the temporal characteristics of the original sequence, ensuring a comprehensive temporal-spatial representation. \(D\) represents the hidden channel size.
\subsubsection{Spike-based Multimodal Fusion Module }
Multimodal skeleton data consists of four components: bones (\(X_{\mathrm{b}}\)), joints (\(X_{\mathrm{j}}\)), bone motions (\(X_{\mathrm{bm}}\)), and joint motions (\(X_{\mathrm{jm}}\)), each represented with dimensions \((C, T, V)\).
The multimodal skeleton data are first represented independently through the SSC module to obtain their respective spike-form features: \(P_{\mathrm{b}}, P_{\mathrm{j}}, P_{\mathrm{bm}}, P_{\mathrm{jm}} \in \mathbb{R}^{S \times D \times V \times T}\) and then fed into the SMF module. The SMF module is based on mutual information for the fusion of sparse spike-form features derived from multimodal data. 

The spike-form features are then pairwise selected and concatenated along the feature dimension \( D \) to construct a joint distribution. To construct a corresponding marginal distribution, one of the two spike-form features in each pair is randomly shuffled along the spike time step dimension \(S\), and the same concatenation is performed.
We propose the Spike-form Mutual Information Calculation (SMIC) module, formally composed of LSTM, SN, FC, and GAP layers, to realize spiking-driven mutual information prediction.
The framework computes the mutual information between four sparse spike-form features and performs weighted fusion. By leveraging the computational efficiency of sparse spike-form features and maximizing the lower bound of mutual information, the SMIC network effectively captures shared information across modalities, facilitating robust and energy-efficient multimodal feature fusion within SNN framework, as illustrated in Figure~\ref{fig:sub2}.

 To illustrate the process, we take bone (\(P_{\mathrm{b}}\)) and joint (\(P_{\mathrm{j}}\)) data as an example. First, the spike-form features from the two modalities are concatenated along the feature dimension to construct a joint distribution:
 \begin{equation}
P_{\mathrm{Joint}} = \operatorname{cat}(P_{\mathrm{j}}, P_{\mathrm{b}}, \operatorname{dim}=1)\in \mathbb{R}^{S \times 2D \times V \times T}.
\end{equation}
To generate a marginal distribution, the spike-form feature \(P_{\mathrm{b}}\) is shuffled randomly along the spike time step dimension (\(S\)), resulting in \(P_{\mathrm{b}}^{\prime}\). The shuffled feature is then concatenated with \(P_{\mathrm{j}}\):
\begin{equation}
P_{\text{marginal}} = \operatorname{cat}(P_{\mathrm{j}}, P_{\mathrm{b}}^{\prime}, \operatorname{dim}=1)\in \mathbb{R}^{S \times 2D \times V \times T}.
\end{equation}
Both \(P_{\mathrm{Joint}}\) and \(P_{\text{marginal}}\) are then processed through an LSTM network equipped with SN layers. For the joint distribution, the output of the LSTM is passed through a FC layer followed by a GAP operation as:
\begin{equation}
\boldsymbol{t}_1 = \operatorname{SN}(\operatorname{LSTM}(P_{\mathrm{Joint}})), \quad \boldsymbol{t} = \operatorname{GAP}(\operatorname{FC}(\boldsymbol{t}_1)).
\end{equation}
Similarly, for the marginal distribution, the output is processed through the same layers, and an additional exponential operation is applied:
\begin{equation}
\boldsymbol{et}_1 = \operatorname{SN}(\operatorname{LSTM}(P_{\text{marginal}})), \quad \boldsymbol{et} = \exp\left(\operatorname{GAP}(\operatorname{FC}(\boldsymbol{et}_1))\right).
\end{equation}
Based on the predictions \(\boldsymbol{t}\) and \(\boldsymbol{et}\), the lower bound of mutual information is estimated as:
\begin{equation}
mi\_lb_{\mathrm{jb}}= \bar{\boldsymbol{t}} - \log \overline{\boldsymbol{et}}.
\end{equation}
The calculation is performed pairwise for all modalities—bones, joints, bone motions, and joint motions—resulting in six pairwise mutual information values, which are organized into a symmetric matrix \(\boldsymbol{MI}\) with zeros along the diagonal:
\begin{equation}
\boldsymbol{MI} = \begin{bmatrix}
0 & \cdots & mi\_lb_{\mathrm{jb}} \\
\vdots & \ddots & \vdots \\
mi\_lb_{\mathrm{bj}} & \cdots & 0
\end{bmatrix}_{4 \times 4}.
\end{equation}
To normalize the mutual information values, the sum along the rows of \(\boldsymbol{MI}\) is computed, yielding the average mutual information weights:
\begin{equation}
\boldsymbol{\overline{MI}} = \frac{\sum_{r=1}^4 \boldsymbol{MI}_{r,:}}{\sum_{r=1}^4 \sum_{c=1}^4 \boldsymbol{MI}_{r,c}},
\end{equation}
\begin{equation}
\boldsymbol{W}_{\text{MI}} = \frac{\boldsymbol{\overline{MI}} - \min(\boldsymbol{\overline{MI}})}{\max(\boldsymbol{\overline{MI}}) - \min(\boldsymbol{\overline{MI}})} \in \mathbb{R}^4.
\end{equation}
The fused spike-form features are then obtained by weighting the original features of all modalities using \(\boldsymbol{W}_{\text{MI}}\):
\begin{equation}
P_{\text{fuse}} = \boldsymbol{W}_{\text{MI}} * \left[P_{\mathrm{b}}, P_{\mathrm{j}}, P_{\mathrm{bm}}, P_{\mathrm{jm}}\right]^{\boldsymbol{T}} \in \mathbb{R}^{S \times D \times V \times T}.
\end{equation}
Finally, the fused feature \(P_{\text{fuse}}\) is fed into the SGN Block for subsequent processing. By leveraging spiking-driven mutual information, the SMF module ensures the effective integration of sparse multimodal features, enhancing the robustness and accuracy of the overall framework.
\subsection{SGN Block}
 Revisiting vanilla skeleton-based action recognition based on GCN, we construct our SGN with spike-form features and adjacency matrices. The process can be summarized as follows:
\begin{equation}
\mathcal{A} = \Delta^{-\frac{1}{2}} \widetilde{A} \Delta^{-\frac{1}{2}}, \quad 
\end{equation}
where \(\mathcal{A}\) is the normalized adjacency matrix of dimension \( K \times V \times V \). \(K\) represents the number of branches, each capturing distinct relationships between nodes, and \(V\) denotes the number of nodes in the graph. \(\widetilde{A} = A + I\) denotes the sum of the adjacency matrix \(A\) and the identity matrix \(I\) (representing self-connections), while \(\Delta\) is the diagonal degree matrix of \(\widetilde{A}\).\\
\indent To calculate the multi-branch graph convolution, the branch-specific adjacency matrix \( \mathcal{A}_k \in \mathbb{R}^{V \times V} \) is applied to the input spiking-form feature \( P \in \mathbb{R}^{S \times D \times T \times V} \). The first layer of the SGC begins by aggregating outputs from all \( K \) branches and is defined as:
\begin{equation}
H^{(0)} = \mathrm{SN}\left(\mathrm{BN}\left(PW_\text{r}^{(0)}\right)\right) + \mathrm{SN}\left(\mathrm{BN}\left(\sum_{k=1}^{K} \mathcal{A}_k P W_k^{(0)}\right)\right),
\end{equation}
where \( \mathcal{A}_k \) represents the adjacency matrix for branch \( k \), \( W_k^{(0)} \text{ and }  W_\text{r} \in \mathbb{R}^{D \times D^{\prime}} \) are the learnable weight matrix specific to the \( k \)-th branch and a residual connection, and \( \mathrm{SN} \) applied to the aggregated features. To further enhance the model's representation capabilities, we incorporate SSA following the principles of existing spiking attention methods\cite{zhou2022}. These methods leverage a spiking-based \( QKV \) structure with scaling factors \( s \) to control large matrix multiplication values. The computation of SSA can be described as follows:
\begin{equation}
Q^{(0)} = \mathrm{SN}\left(\mathrm{BN}\left(H^{(0)} W^{(0)}_Q\right)\right), \quad 
\end{equation}
\begin{equation}
K^{(0)} = \mathrm{SN}\left(\mathrm{BN}\left(H^{(0)} W^{(0)}_K\right)\right), \quad 
\end{equation}
\begin{equation}
V^{(0)} = \mathrm{SN}\left(\mathrm{BN}\left(H^{(0)}W^{(0)}_V\right)\right), \quad 
\end{equation}
\begin{equation}
H^{(0)}_\text{SA} = H^{(0)} + \operatorname{SN}\left((Q^{(0)} {K^{(0)}}^{T}) V^{(0)} \cdot s\right), \quad 
\end{equation}
where, \( Q^{(0)} \), \( K^{(0)} \), and \( V^{(0)} \) are computed through \(\mathrm{SN}\) after \(\mathrm{BN}\) and linear transformations using learnable weights \( W^{(0)}_Q, W^{(0)}_K, \) and \( W^{(0)}_V \). The attention mechanism integrates the weighted interactions, while the scaling factor \( s \) prevents large values in the spiking attention computation. \( H^{(0)}_\text{SA} \) combines the residual connection \( H^{(0)} \) with the attention-enhanced feature, defined as SA-SGC, illustrated in Figure~\ref{fig:MK-SGN}(b). SA-SGC  yields a robust representation for spiking graph processing.
After that, the first layer of STC, illustrated in Figure~\ref{fig:MK-SGN}(c), can be written as:
\begin{equation}
T^{(0)} = \operatorname{SN}\left(\mathrm{BN}\left(W_\text{t}^{(0)}\left(H^{(0)}_\text{SA}\right)\right)\right) + \operatorname{SN}\left(H^{(0)}_\text{SA}\right),
\end{equation}
where $W_\text{t}$ is the learnable weight matrix.
For subsequent layers, the spiking self  attention-enhanced feature \( H^{(l)}_\text{SA} \)  and temporal features \( T^{(l-1)} \), expressed as:
\begin{equation}
H^{(l)} = \mathrm{SN}\left(\mathrm{BN}\left(T^{(l-1)}W_\text{r}^{(l)}\right)\right) + \mathrm{SN}\left(\mathrm{BN}\left(\sum_{k=1}^{K} \mathcal{A}_k T^{(l-1)} W_k^{(l)}\right)\right),
\end{equation}
\begin{equation}
H^{(l)}_\text{SA} = H^{(l)} + \operatorname{SN}\left((Q^{(l)} {K^{(l)}}^{T}) V^{(l)} \cdot s\right), \quad 
\end{equation}
\begin{equation}
T^{(l)} = \operatorname{SN}\left(\mathrm{BN}^{(l)}\left(W_\text{t}^{(l)}\left(H^{(l)}_\text{SA}\right)\right)\right) + \operatorname{SN}\left(H^{(l)}_\text{SA}\right).
\end{equation}
Finally, the output features of the ast layer \(L\) are processed through GAP and a FC layer to generate the prediction, as follows:
\begin{equation}
y = \operatorname{FC}\left(\mathrm{GAP}\left(\operatorname{SN}\left(T^{(L)}\right)\right)\right),
\end{equation}
where \( y \in \mathbb{R}^{U} \), and \( U \) is the number of action classes. We further calculate the standard cross-entropy loss 
 $\mathcal{L}_{\mathrm{task}}$ based on the predicted probabilities and one-hot encoded ground-truth labels.

\subsection{Knowledge Distillation of GCN-to-SGN}
\begin{figure}[t]
    \centering
    \begin{subfigure}[b]{0.48\textwidth}
        \centering
        \includegraphics[width=\textwidth]{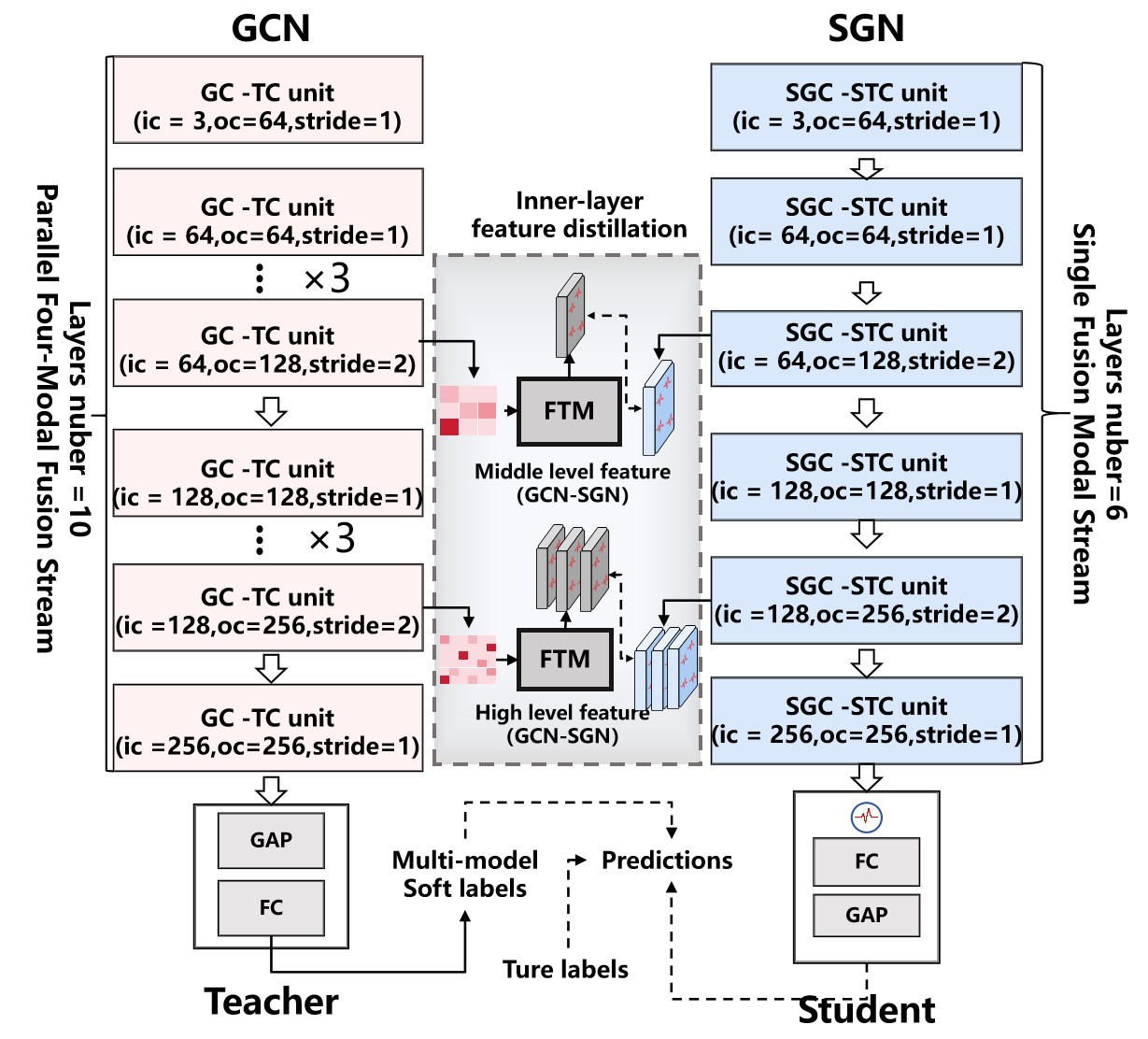}
        \caption{\textcolor{black}{Inner-layer feature distillation and soft label distillation.}}
        \label{fig:ARC}
    \end{subfigure}\hfill
    \begin{subfigure}[b]{0.3\textwidth}
        \centering
        \includegraphics[width=\textwidth]{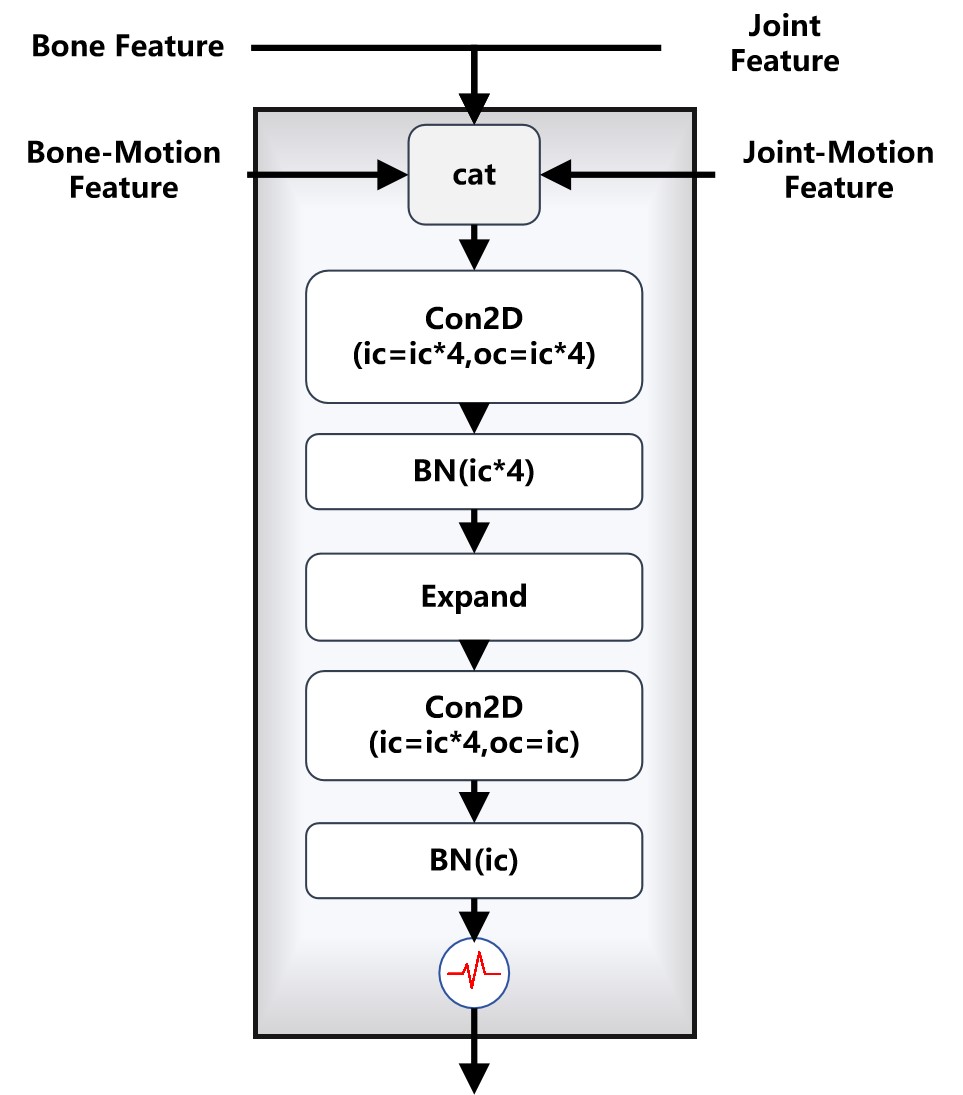}
        \caption{\textcolor{black}{The detail structure of FTM}}
        \label{fig:FT}
    \end{subfigure}
    \caption{\textcolor{black}{GCN-to-SGN Knowledge Distillation Method}}
    \label{fig:ANN-SNN}
\end{figure}

To leverage the powerful representation capabilities of multimodal GCNs, we propose a novel distillation framework with the multimodal GCN serving as the teacher network and the SGN as the student network, shown in Figure~\ref{fig:ANN-SNN}.

The proposed approach fully exploits the rich knowledge embedded in pre-trained multimodal GCNs, transferring superior representational features to the SGN. By aligning the intermediate features and outputs of the SGN with those of the GCN, shown in Figure~\ref{fig:ARC}, the method significantly enhances SGN accuracy. The enhancement is achieved without introducing additional computational complexity or energy consumption during the inference phase, thereby preserving the inherent efficiency advantages of the SGN architecture. The distillation framework bridges the performance gap between GCNs and SGNs while maintaining the suitability of SGNs for energy-efficient applications.
\subsubsection{Soft Label Distillation}
In the proposed distillation framework, the CTR-GCN with a 4-stream architecture is utilized as the teacher network \cite{liu2020}. Multimodal data inputs, including bone, joint, bone-motion, and joint-motion data, are processed independently by single-stream CTR-GCN. Single-modal soft labels, represented as \(y_\text{b}\), \(y_\text{j}\), \(y_\text{bm}\), and \(y_\text{jm}\), are generated corresponding to the respective modalities.

The outputs of the four modalities are then aggregated to form the multimodal soft labels \(y_{\text{mm}}\), which serve as the knowledge to be distilled into the SGN. This aggregation is performed using weighted summation, as described below:
\begin{equation}
y_{\text{mm}} = \alpha_\text{b} y_\text{b} + \alpha_\text{j} y_\text{j} + \alpha_\text{bm} y_\text{bm} + \alpha_\text{jm} y_\text{jm},
\end{equation}
where \(\alpha_\text{b}\), \(\alpha_\text{j}\), \(\alpha_\text{bm}\), and \(\alpha_\text{jm}\) are hyperparameters that control the relative contribution of each modality. These parameters are set based on the importance or reliability of the modalities during training.

The SGN then learns to approximate the multimodal teacher soft labels $y_{\text{mm}}$ through a regression-based loss function. The final learning objective for soft label distillation is expressed as:
\begin{equation}
\mathcal{L}_\text{sdk} = \|y - y_{\text{mm}}\|,
\end{equation}
where \(y\) represents the predictions of the SGN. By aligning its outputs with the teacher's multimodal soft labels, the SGN inherits the representational power of the multimodal GCN, improving its performance without increasing its energy consumption or computational cost.
\subsubsection{Inner-layer feature distillation }
The inner-layer features of heterogeneous GCN and SGN models differ significantly. While GCN typically processes continuous decimal values, SGN operates on sparse discrete spike sequences. To address this discrepancy, we propose the FTM, as illustrated in Figure~\ref{fig:FT}. The FTM is a learnable module that transforms the continuous intermediate features of GCN into spike-form features, making them compatible with SGN.

The FTM concatenates multimodal features along the channel dimension and fuses them using depth-wise separable convolution. Subsequently, continuous features are expanded. Finally, spike-form features are generated through Conv2D, BN, and SN layers. This ensures compatibility between the teacher GCN and student SGN architectures while preserving the valuable structural and semantic information embedded in the GCN features.
In our distillation framework, the teacher GCN is composed of 10 GC-TC (Graph Convolution-Temporal Convolution) units, while the student SGN consists of 6 SGC-STC (Spiking Graph Convolution-Spiking Temporal Convolution) units. For feature distillation, we leverage intermediate-level features (\(T_{\mathrm{A b5}}, T_{\mathrm{A j5}}, T_{\mathrm{A bm5}}, T_{\mathrm{A jm5}}\)) and high-level features (\(T_{\mathrm{A b8}}, T_{\mathrm{A j8}}, T_{\mathrm{A bm8}}, T_{\mathrm{A jm8}}\)). The intermediate features are represented as:
$T_{\mathrm{A b5}}, T_{\mathrm{A j5}}, T_{\mathrm{A bm5}}, T_{\mathrm{A jm5}} \in \mathbb{R}^{2C \times \frac{T}{2} \times V},$ while the high-level features are:$T_{\mathrm{A b8}}, T_{\mathrm{A j8}}, T_{\mathrm{A bm8}}, T_{\mathrm{A jm8}} \in \mathbb{R}^{4C \times \frac{T}{4} \times V}.$
These layers are selected because the expansion of channel dimensions and the changes in stride within the TCN network allows them to capture diverse temporal relationships and rich feature information. Taking the intermediate 5th layer as an example, the multimodal teacher features are first concatenated and fused as follows:
\begin{equation}
T_{\mathrm{A m5}} = \operatorname{BN}\left(\operatorname{Conv2D}\left(\operatorname{cat}\left(T_{\mathrm{A b5}}, T_{\mathrm{A j5}}, T_{\mathrm{A bm5}}, T_{\mathrm{A jm5}}\right)\right)\right),
\end{equation}
where \(T_{\mathrm{A m5}} \in \mathbb{R}^{8C \times \frac{T}{2} \times V}\) represents the fused multimodal teacher features. These features are then translated into spike-form features through the FTM:
\begin{equation}
T_{\mathrm{S m5}} = \operatorname{SN}\left(\operatorname{BN}\left(\operatorname{Conv2D}\left(\operatorname{Expand}\left(T_{\mathrm{A m5}}\right)\right)\right)\right),
\end{equation}
where \(T_{\mathrm{S m5}} \in \mathbb{R}^{S \times D \times V \times \frac{T}{2}}\) are the spike-form features. Similarly, the spike-form features for the high-level 8th layer (\(T_{\mathrm{S m8}}\)) are calculated.\\
\indent To align the student SGN features with the teacher GCN features, we use cosine similarity as the loss function. For the converted spike-form feature from the multimodal GCN and the corresponding 3rd-layer spike-form feature in SGN (\(T^{(3)}\)), the loss is defined as:
\begin{equation}
\mathcal{L}_{\mathrm{fkd}_1} = \frac{1}{J} \sum_{j=1}^{J} \left(1 - \frac{T_{\mathrm{S m5}}^j \cdot {T^{(3)}}^j}{\|T_{\mathrm{S m5}}^j\| \cdot \|{T^{(3)}}^j\|}\right),
\end{equation}
where \(J\) is the number of samples. Similarly, for the high-level 8th-layer spike-form feature and the SGN 5th-layer feature (\(T^{(5)}\)):
\begin{equation}
\mathcal{L}_{\mathrm{fkd}_2} = \frac{1}{J} \sum_{j=1}^{J} \left(1 - \frac{T_{\mathrm{S m8}}^j \cdot {T^{(5)}}^j}{\|T_{\mathrm{S m8}}^j\| \cdot \|{T^{(5)}}^j\|}\right).
\end{equation}
The inner-layer feature distillation loss is then defined as:
\begin{equation}
\mathcal{L}_{\mathrm{fkd}} = \beta_1 \mathcal{L}_{\mathrm{fkd}_1} + \beta_2 \mathcal{L}_{\mathrm{fkd}_2},
\end{equation}
where \(\beta_1\) and \(\beta_2\) are hyperparameters that control the weighting of the intermediate and high-level distillation losses. This formulation ensures that the student network effectively learns from the teacher's intermediate and high-level spike-form features while optimizing the FTM for feature translation.\\
\indent Finally, the total loss function combines the inner-layer feature distillation loss with the task loss \(\mathcal{L}_{\mathrm{task}}\) and the soft label distillation loss \(\mathcal{L}_{\mathrm{sdk}}\) as follows:
\begin{equation}
\mathcal{L} = \gamma_1 \mathcal{L}_{\mathrm{task}} + \gamma_2 \mathcal{L}_{\mathrm{sdk}} + \gamma_3 \mathcal{L}_{\mathrm{fkd}},
\end{equation}
where \(\gamma_1\), \(\gamma_2\), and \(\gamma_3\) are hyperparameters that balance the contribution of each loss term. This unified objective ensures that the SGN effectively inherits the representational power of the GCN while maintaining its energy efficiency and computational advantages.
\section{Experiments}
This chapter presents the implementation details, ablation studies, result comparisons, and visualizations. The details will be described in the following subsections.
\subsection{Experimental Setup}
\subsubsection{Datasets}
We evaluate the proposed method on three widely used skeleton-based action recognition datasets: NTU-RGB+D, NTU-RGB+D 120, and NW-UCLA\cite{shahroudy2016,liu2019ntu,wang2014}. These datasets provide diverse action classes, extensive skeleton data, and varying levels of complexity, making them ideal benchmarks for comprehensive evaluation.

NTU-RGB+D \cite{shahroudy2016} is a large-scale dataset comprising 60 action classes and 56,000 video clips. It includes detailed skeleton data covering various daily activities captured using a Kinect v2 sensor. The dataset provides 3D coordinates of 25 major body joints, enabling precise modeling of human movements. NTU-RGB+D is widely recognized as a standard benchmark in the action recognition field due to its variety of actions and large-scale data volume. The dataset is evaluated under two standard settings: Cross-Subject (XSub), where training and testing samples are divided based on the subjects performing the actions, and Cross-View (XView), where training and testing samples are divided based on the camera views.

    NTU-RGB+D 120 \cite{liu2019ntu}, an extended version of NTU-RGB+D, significantly expands the dataset to include 120 action classes and over 114,000 video clips. This dataset offers richer skeleton data, making it one of the largest and most comprehensive resources for skeleton-based action recognition. Its inclusion of a broader range of actions and more extensive data volume provides an ideal foundation for developing and evaluating advanced action recognition algorithms. Similarly, this dataset is evaluated under two standard settings: Cross-Subject (XSub) and Cross-Setup (XSet), where training and testing samples are divided based on the camera setups used during data collection.

NW-UCLA \cite{wang2014} is a smaller dataset comprising 10 action classes and 1,494 video clips captured from multiple viewpoints. It provides 3D coordinates of 20 major body joints, offering diverse skeleton data for assessing the robustness of action recognition models to viewpoint variations. The dataset poses a challenge for models to perform consistently across different perspectives. The evaluation is conducted under the Cross-View setting, where training and testing samples are divided based on the camera angles used for data capture.

The specific statistics for the three datasets after the splits are summarized in Table~\ref{tab:dataset_details}.
\begin{table}[t]
\centering
\caption{The details of the datasets used in our experiments.}
\label{tab:dataset_details}
\resizebox{0.45\textwidth}{!}{%
\begin{tabular}{ccccccc}
\hline
\multirow{3}{*}{\textbf{Dataset}} & \multirow{3}{*}{\textbf{\# Classes}} & \multicolumn{4}{c}{\textbf{Split}}                                                        & \multirow{3}{*}{\textbf{\# Joints}} \\
                                  &                                      & \multicolumn{2}{c}{\textbf{Cross-Subject}} & \multicolumn{2}{c}{\textbf{Cross-View (XSet)}} &                                  \\
                                  &                                      & \textbf{\# Train}      & \textbf{\# Test}      & \textbf{\# Train}       & \textbf{\# Test}       &                                  \\ \hline
NTU RGB+D\cite{shahroudy2016}                         & 60                                   & 40,320               & 16,560              & 37,920                & 18,960               & 25                               \\
NTU RGB+D 120\cite{liu2019ntu}                     & 120                                  & 63,026               & 53,370              & 54,471                & 61,925               & 25                               \\
NW-UCLA\cite{wang2014}                         & 10                                   & -                    & -                   & 1,039                 & 455                  & 20                               \\ \hline
\end{tabular}%
}
\end{table}
\subsubsection{Implementation Details}
The experiments are conducted on a server equipped with an NVIDIA SMI V100 GPU with 32 GB of memory. The models are implemented using PyTorch\cite{pytorch} and SpikingJelly\cite{spikingjelly}, which provide efficient frameworks for spiking neural networks. 

The key hyperparameters for the LIF model used in all experiments are listed in Table~\ref{4}. The training configurations for MK-SGN across different datasets are summarized in Table~\ref{5}. Additionally, Table~\ref{table:loss_hyperparams} details the hyperparameters for soft label aggregation and loss weighting, including the weights assigned to different modalities (\(\alpha_b, \alpha_j, \alpha_{bm}, \alpha_{jm}\)) and the balancing factors for various loss components (\(\beta_1, \beta_2, \gamma_1, \gamma_2, \gamma_3\)). 
\begin{table}[t]
\centering
\caption{Hyperparameter settings for the LIF model.}
\begin{tabular}{c|c}
\hline
\textbf{Parameter} & \textbf{Value} \\
\hline
Threshold \(V_{th}\) & 1.0 \\
Reset potential \(V_{reset}\) & 0 \\
Decay factor \(\tau\) & 0.25 \\
Surrogate function's window size \(a\) & 1.0 \\
\hline
\end{tabular}
\label{4}
\end{table}
\begin{table}[t]
\centering
\setlength{\tabcolsep}{1.3pt} 
\footnotesize 
\caption{Hyperparameter settings for MK-SGN training.}
\begin{tabular}{c|c|c|c}
\hline
\textbf{Parameter} & \textbf{NTU-RGB+D} & \textbf{NTU-RGB+D} & \textbf{NW-UCLA} \\
\textbf{} & \textbf{} & \textbf{120} & \textbf{} \\
\hline
Learning Rate & 0.1/0.1 & 0.1/0.1 & 0.1/0.1 \\
Decay Rate & 0.1/0.1 & 0.1/0.1 & 0.1/0.1 \\
Batch Size & 64 & 64 & 128 \\
Time Steps & 4 & 4 & 4 \\
Training Epochs & 60 & 60 & 60 \\
Step & 50 & 50 & 50 \\
Dropout Rate & 0.3 & 0.3 & 0.3 \\
Weight Decay & 1e-4 & 1e-4 & 1e-4 \\
Optimizer & SGD & SGD & SGD \\
\hline
\end{tabular}
\label{5}
\end{table}
\begin{table}[t]
\caption{Hyperparameters for soft label aggregation and loss weighting.}
\centering
\setlength{\tabcolsep}{6pt} 
\renewcommand{\arraystretch}{1.2} 
\footnotesize 
\begin{tabular}{lll}
\hline
\textbf{Hyperparameter} & \textbf{Description} & \textbf{Value} \\
\hline
\(\alpha_b\)            & Weight for bone modality.              & 0.25 \\
\(\alpha_j\)            & Weight for joint modality.             & 0.25 \\
\(\alpha_{bm}\)         & Weight for bone motion modality.       & 0.25 \\
\(\alpha_{jm}\)         & Weight for joint motion modality.      & 0.25 \\
\(\beta_1\)             & Weight of intermediate distillation loss. & 0.5  \\
\(\beta_2\)             & Weight of high-level distillation loss.   & 0.5  \\
\(\gamma_1\)            & Weight of task loss term.              & 1.0  \\
\(\gamma_2\)            & Weight of soft label loss term.        & 1.0  \\
\(\gamma_3\)            & Weight of feature distillation loss term. & 1.0  \\
\hline
\end{tabular}
\label{table:loss_hyperparams}
\end{table}
The training configurations for MK-SGN across different datasets are summarized in Table~\ref{5}. We adopt an initial learning rate of 0.1, a weight decay rate of 1e-4, and set the batch size based on the dataset: 64 for NTU-RGB+D, 64 for NTU-RGB+D 120, and 128 for NW-UCLA. Training is conducted for 64 epochs with a step decay schedule, reducing the learning rate to one-tenth after 50 epochs. In the testing phase, only the SGN model is evaluated. This ensures that the SGN's performance reflects its learned capabilities without additional computational overhead.

\subsection{Ablation Studies}
To verify the effectiveness and efficiency of the proposed model, we first replace the activation layers of the classic one-stream 2S-AGCN model \cite{shi2019} with LIF neurons. This modified model, called Base-SGN*, is then trained and tested for comparison. Additionally, we compare our proposed model with Spikformer\cite{zhou2022}, a classic SNN-based transformer architecture. For a fair comparison, we adjust the layer structure and hidden dimension of Spikformer to match the MK-SGN architecture. These adjustments ensure that the comparison focuses on the modeling approach while minimizing the influence of structural differences. Based on these baselines, we conduct ablation and comparison experiments to evaluate the performance of MK-SGN.
\subsubsection{Theoretical Energy Consumption Calculation}
In the context of SNNs, synaptic operations (\(\operatorname{SOPs}\)) refer to the fundamental operations required to transmit spikes across synapses and perform computations within spiking neural networks. Specifically, \(\operatorname{SOPs}\) are analogous to \(\operatorname{FLOPs}\) in ANNs, serving as the core unit of the computational workload. 

Each \(\operatorname{SOP}\) corresponds to a spike-based accumulate (\(\operatorname{AC}\)) operation, which involves multiplying a binary spike (0 or 1) by a synaptic weight and accumulating the result in the post-synaptic neuron. This operation leverages the simplicity of binary spike-based computations, significantly reducing computational complexity compared to ANNs. In ANNs, operations are defined as Multiply-and-Accumulate (\(\operatorname{MACs}\)), where both multiplication and addition involve floating-point numbers. The use of \(\operatorname{SOPs}\) in spiking neural networks thus highlights their efficiency, as \(\operatorname{ACs}\) require fewer computational resources than \(\operatorname{MACs}\).

As one of the critical components in our experimental evaluation, we calculate the SOPs for both Base-SGN* and MK-SGN to achieve two key objectives. First, this analysis underscores the advantages of our model on neuromorphic hardware by demonstrating its computational efficiency. Second, the calculated SOPs serve as the basis for estimating theoretical energy consumption, providing evidence of the energy-saving potential of MK-SGN. 

\begin{table*}[t]
\caption{\textcolor{black}{Evaluation on NTU-RGB+D. Param refers to the number of parameters. Power is the average theoretical energy consumption when predicting a skeleton from the NTU-RGB+D test set, whose calculation details are shown in Equation(35), Equation(36), and Equation(37).}}
\centering
\setlength{\tabcolsep}{5pt} 
\footnotesize 
\resizebox{\textwidth}{!}{%
\begin{threeparttable}
\begin{tabular}{ccccccccccl}
\hline
\multirow{2}{*}{\textbf{Method}} & \multirow{2}{*}{\textbf{Architecture}} & \textbf{Param} & \multirow{2}{*}{\textbf{\begin{tabular}[c]{@{}c@{}}Spike Time\\ Step\end{tabular}}} & \multirow{2}{*}{\textbf{SMF}}& \multirow{2}{*}{\textbf{\begin{tabular}[c]{@{}c@{}}Knowledge\\ Distillation\end{tabular}}} & \textbf{SOPs}  & \textbf{FLOPs}  & \textbf{Power} & \multicolumn{2}{c}{\textbf{Ntu-RGB+D}} \\
                                 &                                        & \textbf{(M)}   &                                                                               &                                                                                  &                                                                                            & \textbf{(G)} & \textbf{(G)} & \textbf{(mJ)}  & \textbf{Xsub(\%)} & \textbf{Xview(\%)} \\ \hline
ANN                    & (10-layer)ST-GCN \cite{yan2018}                & 3.1           & /                                                                             & /                                                                                & /                                                                                          & / & 3.48        & 16.01          & 81.5          & 88.3          \\ \hline
SNN                       
                                 &(10-layer)Base-SGN*             & 2.07           & 4                                                                             & /                                                                                & /                                                                                          & 0.60   & 7.17       &  0.536                 & 64.2              & 71.3              \\
                                 & (6-layer)Spikformer\cite{zhou2022}\textsuperscript{ICLR 2023}             & 4.78           & 4                                                                             & /                                                                               & /                                                                                          & 1.69   & 24.07       &  2.17                 & 73.9              & 80.6              \\
                                 & \textcolor{black}{(10-layer) MK-SGN} & 
\textcolor{black}{3.13} & 
\textcolor{black}{4} & 
\textcolor{black}{/} & 
\textcolor{black}{/} & 
\textcolor{black}{0.96} & 
\textcolor{black}{10.43} & 
\textcolor{black}{0.86} & 
\textcolor{black}{72.9} & 
\textcolor{black}{79.6} \\
&
\textcolor{black}{(10-layer) MK-SGN} & 
\textcolor{black}{3.13} & 
\textcolor{black}{4} & 
\textcolor{black}{/} & 
\textcolor{black}{\checkmark} & 
\textcolor{black}{0.98} & 
\textcolor{black}{11.34} & 
\textcolor{black}{0.878} & 
\textcolor{black}{74.3} & 
\textcolor{black}{81.2} \\
&
\textcolor{black}{(10-layer) MK-SGN} & 
\textcolor{black}{3.24} & 
\textcolor{black}{4} & 
\textcolor{black}{\checkmark} & 
\textcolor{black}{/} & 
\textcolor{black}{1.00} & 
\textcolor{black}{11.45} & 
\textcolor{black}{0.912} & 
\textcolor{black}{74.2} & 
\textcolor{black}{80.1} \\
&
\textcolor{black}{(10-layer) MK-SGN} & 
\textcolor{black}{3.24} & 
\textcolor{black}{4} & 
\textcolor{black}{\checkmark} & 
\textcolor{black}{\checkmark} & 
\textcolor{black}{1.02} & 
\textcolor{black}{11.87} & 
\textcolor{black}{0.931} & 
\textcolor{black}{78.2} & 
\textcolor{black}{85.1} \\

                                 & (6-layer) MK-SGN              & 1.89           & 4                                                                             & /                                                                                & /                                                                                          & 0.64   & 7.32       &  0.573                 & 74.5              & 80.9              \\
                                 
                                 & (6-layer) MK-SGN              & 1.89           & 4                                                                             & /                                                                                & \checkmark                                                                                  & 0.64  & 7.32        & 0.573                 & 75.9              & 82.1              \\
                                 & (6-layer) MK-SGN              & 2.17           & 4                                                                             & \checkmark                                                                                & /                                                                                  & 0.67   & 7.76       & 0.596                & 76.1          & 82.6          \\

                                 \rowcolor[gray]{0.9}
                                 & (6-layer) MK-SGN              & 2.17           & 4                                                                             & \checkmark                                                                                & \checkmark                                                                                  & 0.67  & 7.76  & 0.596                 & 78.5          & 85.6    \\
&\textcolor{black}{(4-layer) MK-SGN} & 
\textcolor{black}{1.17} & 
\textcolor{black}{4} & 
\textcolor{black}{\checkmark} & 
\textcolor{black}{\checkmark} & 
\textcolor{black}{0.37} & 
\textcolor{black}{4.13} & 
\textcolor{black}{0.212} & 
\textcolor{black}{68.5} & 
\textcolor{black}{71.6} \\
     
                                 \hline
\end{tabular}
\begin{tablenotes}
\item \textit{Note:} FLOPs are calculated using the \href{https://github.com/facebookresearch/fvcore}{FVCORE} library, which provides precise computation of floating-point operations.
\end{tablenotes}
\end{threeparttable}}
\label{ntu}
\end{table*}

Following \cite{zhou2022, yao2024a, yao2024b}, the \(\operatorname{SOPs}\) for a given block or layer \(l\) are defined as the product of the firing rate \(r\), the number of spike time steps \(S\), and the (\(\operatorname{FLOPs}\)) of the block or layer:
\begin{equation}
\operatorname{SOPs}(l) = r \times S \times \operatorname{FLOPs}(l),
\end{equation}
The theoretical energy consumption of Base-SGN and MK-SGN is calculated as follows:
\begin{equation}
\begin{aligned}
E_{\text{Base-SGN}} &= E_{\text{MAC}} \times \mathrm{FL}_{\mathrm{SNN Conv}}^1 \\
&\quad + E_{\text{AC}} \times \left( \sum_{n=2}^N \mathrm{SOP}_{\mathrm{SNN Conv}}^n + \sum_{m=1}^M \mathrm{SOP}_{\mathrm{SNN FC}}^m \right),
\end{aligned}
\end{equation}
\begin{equation}
\begin{aligned}
E_{\text{MK-SGN}} &= n_m \times E_{\text{MAC}} \times \mathrm{FL}_{\mathrm{SNN Conv}}^1 \\
&\quad + E_{\text{AC}} \times \bigg( \frac{k \times (k-1)}{2} \mathrm{SOP}_\mathrm{SMIC} + \sum_{n=n_m}^N \mathrm{SOP}_{\mathrm{SNN Conv}}^n \\
&\quad \quad + \sum_{m=1}^M \mathrm{SOP}_{\mathrm{SNN FC}}^m  + \sum_{l=1}^L \mathrm{SOP}_{\mathrm{SSA}}^l \bigg).
\end{aligned}
\end{equation}

In these equations, \(\mathrm{FL}_{\mathrm{SNN Conv}}^1\) represents the initial layer responsible for encoding skeleton data into spike-form, while \(\mathrm{SOP}_{SMIC}\) accounts for the energy cost of modality fusion with \(k\) denoting the number of modalities.

Energy consumption is estimated under the assumption of 45nm hardware, with \(E_{\text{MAC}} = 4.6 \, \text{pJ}\) and \(E_{\text{AC}} = 0.9 \, \text{pJ}\) \cite{horowitz2014}. The power consumption for a block \(b\) in ANN (MAC-based) and SNN (SOP-based) frameworks is given by:
\begin{equation}
\label{equation:power}
\operatorname{Power}(b) = 4.6 \, \text{pJ} \times \operatorname{FLOPs}(b),
\end{equation}
\begin{equation}
\operatorname{Power}(b) = 0.9 \, \text{pJ} \times \operatorname{SOPs}(b).
\end{equation}

These formulations demonstrate how MK-SGN balances high performance with reduced theoretical energy consumption. 

\subsubsection{The Effect of Different Components}
We use theoretical formulas to estimate the number of operations (OPs) required for classifying a single action sample to analyse the proposed MK-SGN's energy efficiency and evaluate each component's contribution within the model. Specifically, we calculate the FLOPs for ANN-based models, the FLOPs for SNN on traditional GPUs, and the theoretical SOPs on neuromorphic hardware. Based on these operation counts, we further compute the energy consumption per action sample during the inference phase. 

Additionally, we evaluate the classification accuracy of MK-SGN through comprehensive ablation studies conducted on the NTU-RGB+D dataset. The results, including energy efficiency and accuracy comparisons, are summarized in Table \ref{ntu}.

\paragraph{\textbf{Base-SGN*}}
First, we validate the effectiveness of Base-SGN* as the basis for the ablation study. We compare it to the vanilla ST-GCN \cite{yan2018}, calculating the theoretical number of operations (OPs) and energy consumption. The results show that Base-SGN* reduces the theoretical energy consumption to 0.536$mJ$ per action sample. However, this increases the FLOPs and decreases the accuracy to $64.2$ and $71.3$. This indicates that simply replacing the activation layer is insufficient to maintain high classification accuracy.
\paragraph{\textcolor{black}{\textbf{MK-SGN}}}
\textcolor{black}{
Building on Base-SGN*, we compare its performance with the proposed MK-SGN. Even without knowledge distillation, the single-modality MK-SGN achieves significant accuracy improvements of 10.3\% on XSub and 9.6\% on XView, with only a slight increase in OPs and theoretical power consumption. As shown in Table~\ref{ntu}, the 10-layer MK-SGN achieves 72.9\% accuracy on XSub and 79.6\% on XView, with FLOPs at 10.43~G and power consumption at 0.86~mJ. Compared to ST-GCN, MK-SGN offers a better balance between accuracy and computational efficiency, significantly reducing energy consumption.}

\textcolor{black}{Next, we introduce the SMF module and knowledge distillation to further enhance performance. The 10-layer MK-SGN with SMF and knowledge distillation achieves 78.2\% accuracy on XSub and 85.1\% on XView, showcasing the benefits of SMF in improving accuracy while keeping SOPs at 1.02~G and power consumption at 0.931~mJ. Knowledge distillation further boosts accuracy with negligible impact on computational cost.}

\textcolor{black}{For the 6-layer MK-SGN, which eliminates redundant layers from the 10-layer architecture, the SMF and knowledge distillation configurations lead to an accuracy of 78.5\% on XSub and 85.6\% on XView. This version achieves low power consumption of 0.596~mJ and FLOPs of 7.76~G, demonstrating that the 6-layer model is the optimal design for balancing performance and efficiency.}

\textcolor{black}{The 4-layer configuration focuses on further reducing complexity by retaining only the most critical layers (such as channel upsampling layers). This 4-layer MK-SGN achieves lower accuracy—68.5\% on XSub and 71.6\% on XView—but still benefits from the SMF and knowledge distillation techniques, achieving improvements compared to the base model. However, the performance gain is not as substantial as in the 6-layer design, confirming that excessive layer reduction compromises the model’s representation capacity.}

\textcolor{black}{In summary, the addition of SMF and knowledge distillation enhances the accuracy of the model across all configurations. Among the tested architectures, the 6-layer MK-SGN strikes the best balance between accuracy and computational efficiency, demonstrating its suitability for real-time applications that require both high performance and low energy consumption.}

\begin{table}[t]
\caption{Performance comparison for different modality combinations using the SMF module. Accuracy is tested on NTU-RGB+D XSub}
\centering
\setlength{\tabcolsep}{5pt} 
\footnotesize 
\begin{tabular}{cccccc}
\hline
\textbf{Architecture} & \textbf{Modalities} & \textbf{Parameters} & \textbf{Ops} & \textbf{Power} & \textbf{Accuracy} \\
 &  & (M) & (G) & (mJ) & (\%) \\
\hline
MK-SGN               & J               & 1.89                     & 0.64             & 0.573               & 74.5                  \\
                & J+B         & 1.98                     & 0.65             & 0.584               & 75.6                  \\
                & J+B+JM   & 2.08                     & 0.66             & 0.591               & 75.9                  \\
               & J+B+JM+BM & 2.17  & 0.67             & 0.596               & 76.1                  \\
\hline
\end{tabular}
\label{table:smf_results}
\end{table}
\begin{table}[t]
\caption{\textcolor{black}{Comparison of different fusion methods for MK-SGN. Accuracy and theoretical energy consumption are tested on NTU-RGB+D XSub.}}
\centering
\setlength{\tabcolsep}{5pt} 
\footnotesize 

\begin{threeparttable}  

\begin{tabular}{ccccc}
\hline
\textbf{\textcolor{black}{Fusion Method}} & \textbf{\textcolor{black}{Parameters}} & \textbf{\textcolor{black}{Ops}} & \textbf{\textcolor{black}{Power}} & \textbf{\textcolor{black}{Accuracy}} \\
                       &\textcolor{black}{(M)}                &\textcolor{black}{(G)}          & \textcolor{black}{(mJ)}           &\textcolor{black}{(\%)}              \\
\hline
\rowcolor[gray]{0.9}
\textcolor{black}{SMF}                   & \textcolor{black}{2.17}               & \textcolor{black}{0.67}         & \textcolor{black}{0.596}          &\textcolor{black}{76.1}              \\
\textcolor{black}{Spiking Attention}      &\textcolor{black}{2.24}               &\textcolor{black}{0.71}         &\textcolor{black}{0.613}          &\textcolor{black}{76.0}              \\
\textcolor{black}{Concatenation (cat)}    &\textcolor{black}{1.90}              &\textcolor{black}{0.64} &\textcolor{black}{0.589}          &\textcolor{black}{75.4}              \\
\textcolor{black}{Sum (direct addition)}  &\textcolor{black}{1.89}               &\textcolor{black}{0.63}         &\textcolor{black}{0.579}          &\textcolor{black}{75.2}              \\
\hline
\end{tabular}

\begin{tablenotes}
\item \textcolor{black}{\textit{Note:}cat is performed along the channel dimension with downsampling. Spiking Attention involves pairwise modality interaction followed by summation.}
\end{tablenotes}

\end{threeparttable}  

\label{table:fusion_comparison}
\end{table}

\paragraph{\textbf{SMF}}
In multi-modal classification, late fusion requires independent inference for each modality, causing energy consumption to increase by a factor of $n$ for $n$ modalities. For example, fusing four modalities—Joint (J), Bone (B), Joint Motion (JM), and Bone Motion (BM) in ST-GCN~\cite{yan2018}—increases operations to $4 \times 3.48 \, \text{G}$ and power consumption to $4 \times 16.01 \, \text{mJ}$.

To reduce this energy cost while maintaining accuracy, we introduce the SMF module, which integrates modalities more efficiently. Table~\ref{table:smf_results} shows the results from incremental fusion of modalities: J, J+B, J+B+JM, and J+B+JM+BM.

For the single Joint modality, the accuracy is $74.5\%$ with $0.573 \, \text{mJ}$ of power consumption. As more modalities are added, accuracy increases to $76.1\%$ with a slight rise in energy consumption to $0.596 \, \text{mJ}$. The number of parameters increases from $1.89 \, \text{M}$ to $2.17 \, \text{M}$, and operations from $0.64 \, \text{G}$ to $0.67 \, \text{G}$, showing that the SMF module efficiently integrates modalities without significantly increasing complexity.

These results confirm that the SMF module avoids the linear increase in energy consumption associated with late fusion, while improving accuracy by $1.6\%$ compared to single-modality fusion. This demonstrates the SMF module's effectiveness in integrating multiple modalities efficiently.

\paragraph{\textcolor{black}{\textbf{Comparison with Other Fusion Methods}}}

\textcolor{black}{To further assess the effectiveness of the SMF module, we compare it with simpler fusion methods, including Spiking Attention, Concatenation (cat), and direct sum (Sum), using the same four modalities: Joint (J), Bone (B), Joint Motion (JM), and Bone Motion (BM). These methods are commonly used in multimodal fusion tasks but often suffer from higher computational overhead.}

\textcolor{black}{The comparison results are summarized in Table~\ref{table:fusion_comparison}. Spiking Attention and Concatenation (cat) show similar accuracy improvements compared to single-modality fusion, with accuracies of $75.4\%$ and $76.0\%$, respectively. However, both methods require more computational resources than SMF. The Sum method, which directly adds the modality features, results in the lowest accuracy of $75.2\%$, as it fails to effectively model the interactions between modalities.}

\textcolor{black}{The SMF module achieves the highest accuracy of $76.1\%$ with relatively low increase in power consumption ($0.596 \, \text{mJ}$), making it the most efficient method in terms of both accuracy and energy consumption. This validates the superiority of SMF over traditional late fusion techniques, as it balances accuracy improvements with computational efficiency.}

\paragraph{\textbf{Knowledge Distillation of GCN-to-SGN}}

We evaluate the impact of our proposed knowledge distillation method by comparing SGN models trained with and without teacher guidance. The distillation method consists of Soft Label Distillation (SKD) and Inner-layer Feature Distillation (IFD). Ablation experiments are conducted to assess the contributions of each component. Additionally, we examine the effect of SKD using different modality combinations and the influence of IFD at various levels (low-level, high-level, or both).

The results in Table~\ref{table:kd_results} demonstrate the effectiveness of SKD and IFD. With SKD, accuracy improves as more modalities are incorporated: single-modality achieves 75.1\%, two-modalities 75.4\%, and four-modalities 75.6\%, highlighting the benefit of multimodal knowledge transfer. For IFD, focusing on high-level features yields better performance (75.3\%) than low-level features (74.9\%), indicating the significance of abstract information in guiding the student model.

Combining both SKD and IFD in the Full Knowledge Distillation (Full KD) setup achieves 75.9\% accuracy, demonstrating the advantage of integrating soft-label guidance with feature-level distillation.

These results emphasize the importance of combining output-level and feature-level distillation to enhance SGN performance, with SKD aligning the student’s outputs with the teacher’s predictions, and IFD enabling the student to better emulate the teacher’s intermediate representations.
\begin{table}[t]
\caption{Accuracy comparison under different knowledge distillation settings. Accuracy is tested on NTU-RGB+D XSub}
\centering
\resizebox{\linewidth}{!}{
\begin{tabular}{ccccc}
\hline
\textbf{Setting}         & \textbf{SKD} & \textbf{IFD} & \textbf{Modalities} & \textbf{Accuracy (\%)} \\
\hline
-                & -            & -            & -                   & 74.5                  \\
Soft Label (Single Modal.)          & \checkmark   & -            & J              & 75.1                  \\
Soft Label (Two Modal.)  & \checkmark   & -            & J+B             & 75.4                  \\
Soft Label (Four Modal.) & \checkmark   & -            & J+B+JM+BM               & 75.6                  \\
Inner-layer (Low)        & -            & \checkmark   & J+B+JM+BM                   & 74.9                  \\
Inner-layer (High)       & -            & \checkmark   & J+B+JM+BM                   & 75.3                  \\
\rowcolor[gray]{0.9}
Full KD                  & \checkmark   & \checkmark   & J+B+JM+BM                & 75.9                 \\
\hline
\end{tabular}
}
\label{table:kd_results}
\end{table}
\begin{table}[t]
\centering
\setlength{\tabcolsep}{5pt} 
\renewcommand{\arraystretch}{1.2} 
\footnotesize 
\caption{\textcolor{black}{Sensitivity Analysis of Model Hyperparameters and Accuracy for MK-SGN.}}
\textcolor{black}{%
\begin{tabular}{cccccccccc}
\hline
\multicolumn{9}{c}{\textbf{Model Hyperparameters}} & \multirow{2}{*}{\textbf{Accuracy (\%)}} \\
\(\alpha_b\) & \(\alpha_j\) & \(\alpha_{bm}\) & \(\alpha_{jm}\) & \(\beta_1\) & \(\beta_2\) & \(\gamma_1\) & \(\gamma_2\) & \(\gamma_3\) &  \\ \hline
0.1          & 0.1          & 0.1             & 0.1             & 0.5         & 0.5         & 0.5         & 0.5         & 0.5         & 77.1 \\ 
0.5          & 0.5          & 0.5             & 0.5             & 0.5         & 0.5         & 1.0         & 1.0         & 1.0         & 77.5 \\ 
0.25         & 0.5          & 0.25            & 0.25            & 0.5         & 0.5         & 0.8         & 0.9         & 1.0         & 78.3 \\ 
0.75         & 0.25         & 0.75            & 0.75            & 0.5         & 0.5         & 0.6         & 0.7         & 1.0         & 77.9 \\
0.25         & 0.25         & 0.5             & 0.5             & 0.5         & 0.5         & 0.9         & 1.0         & 1.0         & 78.0 \\
0.25         & 0.25         & 0.25            & 0.25            & 0.5         & 0.5         & 1.0         & 1.0         & 1.0         & 78.5 \\ 
0.25         & 0.25         & 0.25            & 0.25            & 1.0         & 1.0         & 0.7         & 0.6         & 0.5         & 77.6 \\ 
0.1          & 0.25         & 0.25            & 0.25            & 1.0         & 0.5         & 0.8         & 0.9         & 0.6         & 77.5 \\ \hline
\end{tabular}
} 
\label{table:model_hyperparams}
\end{table}

\paragraph{\textbf{Integrated Module Performance Comparison}}
We carefully evaluate the contributions of each component in the MK-SGN model, with results summarized in Table~\ref{ntu}. The fully integrated MK-SGN achieves accuracy of 78.5\% and 85.6\% on XSub and XView settings, respectively, with a spike time step of 4. Compared to Spikformer, MK-SGN reduces the parameter count by 54.6\% and FLOPs by 67.7 \%, significantly enhancing its feasibility for training on traditional GPUs. Moreover, MK-SGN decreases SOPs by 60.35\% and theoretical energy consumption by 72.53\% while achieving accuracy improvements of 4.6\% and 5.0\%, demonstrating its ability to balance high performance with low energy consumption.

Compared to Base-SGN*, MK-SGN delivers substantial accuracy gains of 14.3\% and 11.3\% on XSub and XView, respectively. Additionally, when compared to the classic ANN-based model ST-GCN, MK-SGN achieves a remarkable 96.28\% reduction in energy consumption while maintaining comparable accuracy. These results highlight the efficiency and effectiveness of MK-SGN, showcasing its suitability for energy-constrained applications without compromising performance.

\textcolor{black}{%
\paragraph{\textbf{Sensitivity Analysis Experiments}}}

\textcolor{black}{After evaluating the effect of different components in MK-SGN, we conduct sensitivity analysis on key model hyperparameters to assess their impact on accuracy. Unlike training hyperparameters, which have predictable effects, model parameters such as \(\alpha\), \(\beta\), and \(\gamma\) directly influence model performance.}

\textcolor{black}{Experiments are conducted on the NTU-RGB+D XSub dataset. As shown in Table~\ref{table:model_hyperparams}, different configurations of the hyperparameters are tested. The best performance, achieving 78.5\% accuracy, results from a balanced configuration (\(\alpha_b = 0.25, \alpha_j = 0.25, \alpha_{bm} = 0.25, \alpha_{jm} = 0.25\), \(\beta_1 = \beta_2 = 0.25\) and \(\gamma_1 = \gamma_2 = \gamma_3 = 1.0\)).}

\textcolor{black}{The results demonstrate that MK-SGN's accuracy is sensitive to these model hyperparameters. Variations from the optimal configuration lead to a decrease in accuracy, highlighting the importance of tuning these parameters for optimal performance.}
{\color{black}
\subsection{Visualization Analysis}
\paragraph{\textbf{Spiking Self-Attention  Visualization}}
To elucidate how MK-SGN makes decisions and to quantify the contribution of spiking attention across depth and time, we visualize the spike self-attention maps produced by the SSA module, in Fig.\ref{fig:attn_drop_hopping}. Concretely, we render the aggregated \(k\!\odot\!v\) maps (summed over heads) at each layer and along the temporal axis; higher grayscale values indicate stronger spike-modulated attention. The horizontal axis indexes the NTU 25 joints, and the vertical axis indexes time steps \(T\).
\begin{figure*}[t]
\centering
 \begin{subfigure}{1\textwidth}
    \centering
    \includegraphics[width=\linewidth]{attention_map_4.jpg}
    \caption{\textcolor{black}{\textbf{Spiking attention maps for \emph{Drop}.}}}
    \label{fig:attn_drop}
\end{subfigure}
\hfill
\begin{subfigure}{1\textwidth}
    \centering
    \includegraphics[width=\linewidth]{attention_map_17.jpg}
    \caption{\textcolor{black}{\textbf{Spiking attention maps for \emph{Put on Glasses}.}}}
    \label{fig:attn_glasses}
\end{subfigure}
\hfill
\begin{subfigure}{1\textwidth}
    \centering
    \includegraphics[width=\linewidth]{attention_map_25.jpg}
    \caption{\textcolor{black}{\textbf{Spiking attention maps for \emph{Hopping}.}}}
    \label{fig:attn_hopping}
\end{subfigure}
\caption{\textcolor{black}{\textbf{Spiking attention maps for different actions.} The x-axis represents NTU 25 joints, and the y-axis corresponds to time steps \(T\). The grayscale intensity indicates the activation level of the attention mechanism at each joint over time.}}
\label{fig:attn_drop_hopping}
\end{figure*}

\textit{Case A: Drop.}
In Fig.~\ref{fig:attn_drop}, early layers (L1–L2) exhibit diffuse activations. From L3, attention concentrates on the upper-limb chain (shoulder–elbow–wrist–hand, joints 5–12), which directly governs the dropping motion. Moderate responses on the spine chain (joints 1–3, 21) indicate posture stabilization, while the lower-limb chains (joints 13–20) remain weak. In L6, activity condenses into sparse temporal bands aligned with release and follow-through, showing SSA’s ability to emphasize task-critical skeletal substructures.

\textit{Case B: Put on Glasses.}
In Fig.~\ref{fig:attn_glasses}, initial activations are diffuse. From L3, attention intensifies on the hand–arm chains (shoulder–elbow–wrist–hand, joints 5–8, 9–12) and their connection to the head chain (neck–head, joints 3–4, 21). This highlights the coordinated interaction of arms and head required to lift and place glasses. In later layers, spikes are temporally localized around approach and placement, emphasizing fine-grained motor control in upper-limb–head coordination.

\textit{Case C: Hopping.}
As shown in Fig.~\ref{fig:attn_hopping}, early layers (L1–L2) present distributed activations. From L3, focus shifts toward the lower-limb chains (hip–knee–ankle–foot, joints 13–20), which dominate propulsion and landing. In deeper layers, the spine chain (joints 1–3, 21) gains attention, reflecting its stabilizing role during mid-air balance. Final layers show temporally sharp peaks near takeoff and landing, matching the energy-intensive phases of hopping.

This analysis reveals how the attention mechanism evolves to prioritize different joints depending on the action, with the model focusing more on the relevant joints at each phase of the action.

\paragraph{\textbf{Visualization of CAMs on Temporal and Spatial Dimensions}}
To demonstrate the MK-SGN model's effectiveness in capturing key features, we apply Class Activation Mapping (CAM) to the outputs of the SGC and STC layers, as shown in Fig.\ref{fig:cam_visualizations}. The spatial CAM highlights activations from the SGC layer, focusing on relevant joints, while the temporal CAM reflects activations from the STC layer, emphasizing critical time steps. This visualization clearly shows how the model attends to the most important body parts and action phases for accurate classification.
\begin{figure}[t]
\centering
\subfloat[\textbf{\textcolor{black}{Drop}}]{
    \includegraphics[width=0.32\linewidth]{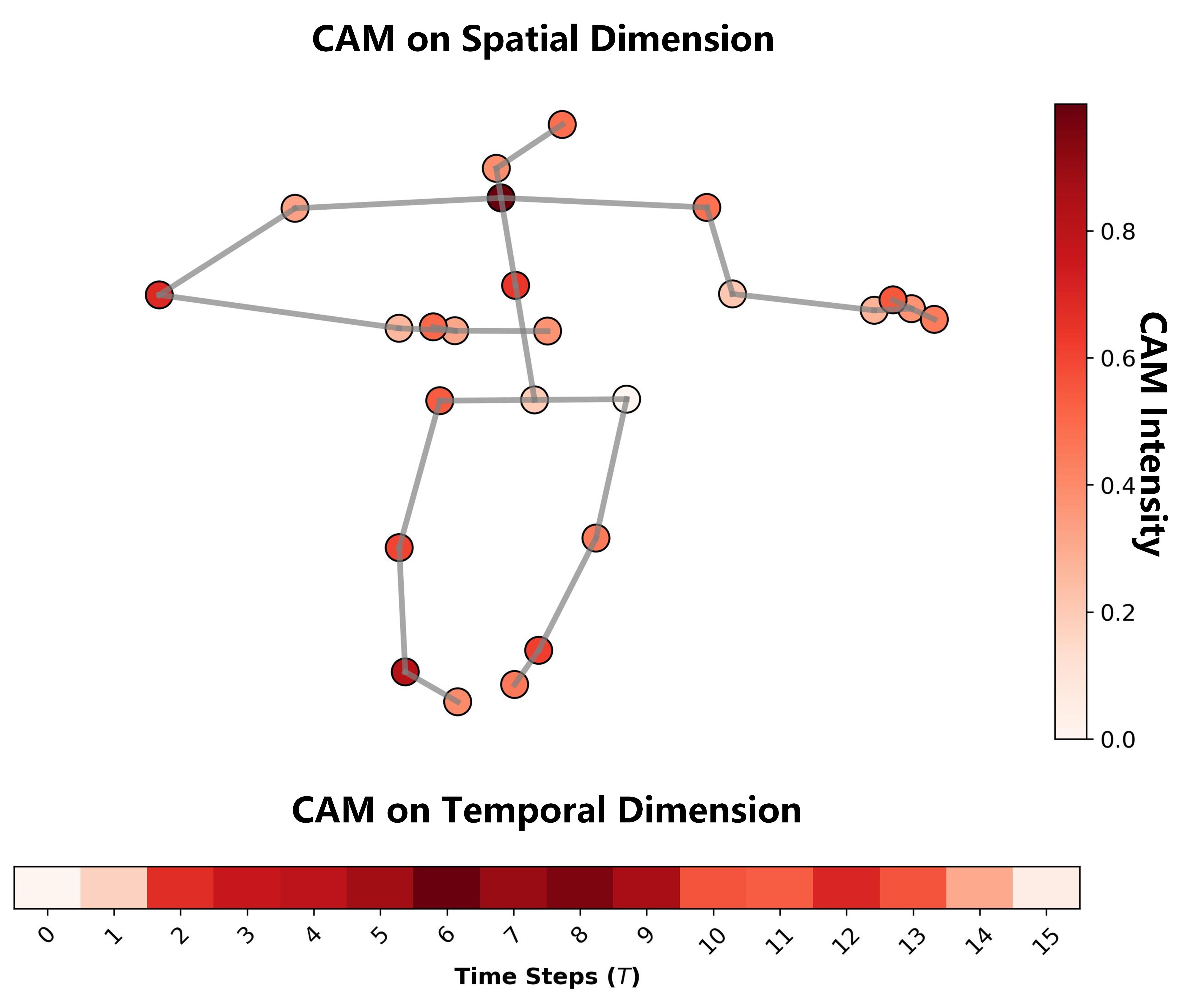}
    \label{fig:cam_4}
}
\subfloat[\textbf{\textcolor{black}{Put on Glasses}}]{
    \includegraphics[width=0.32\linewidth]{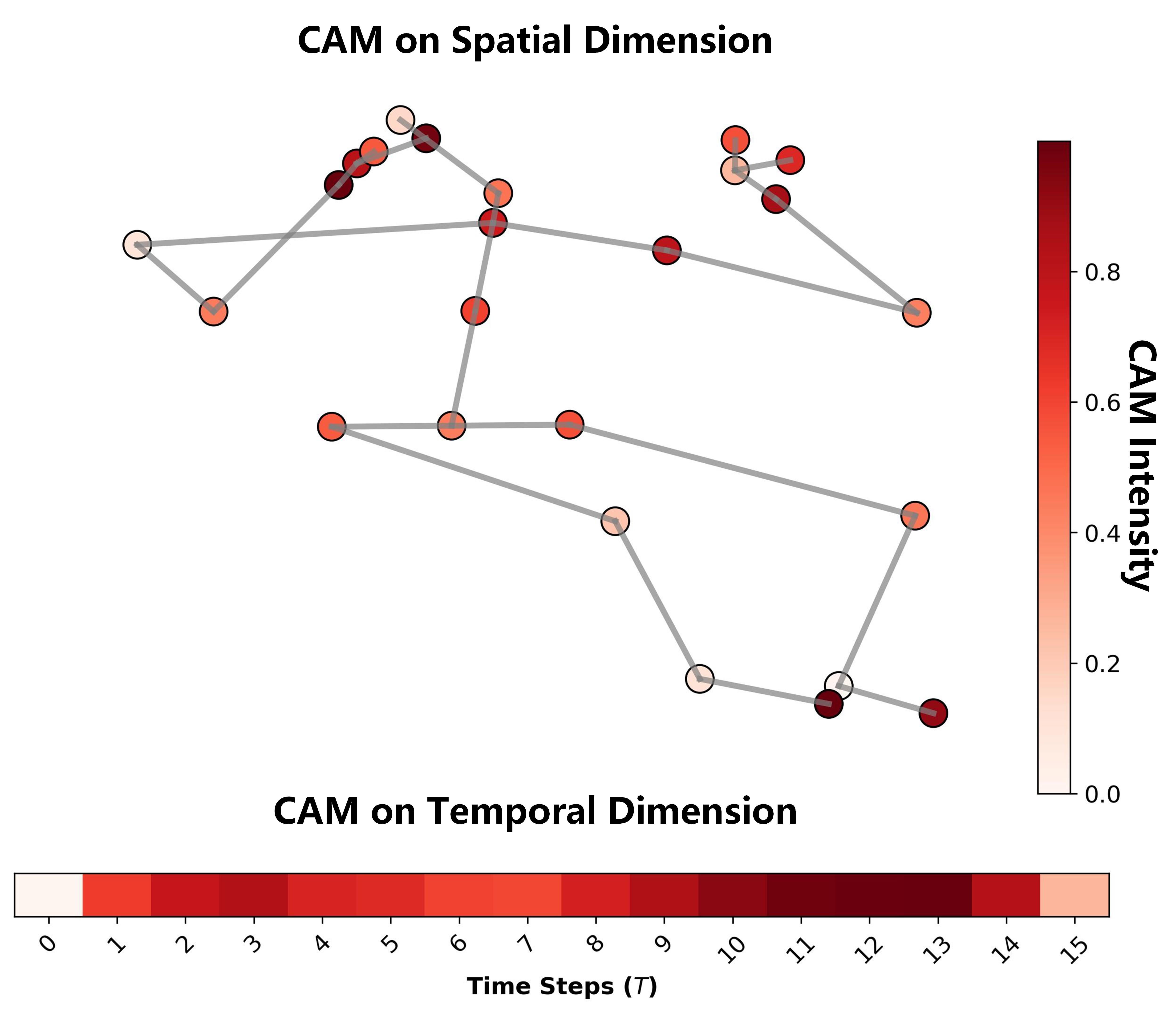}
    \label{fig:cam_17}
}
\subfloat[\textbf{\textcolor{black}{Hopping}}]{
    \includegraphics[width=0.32\linewidth]{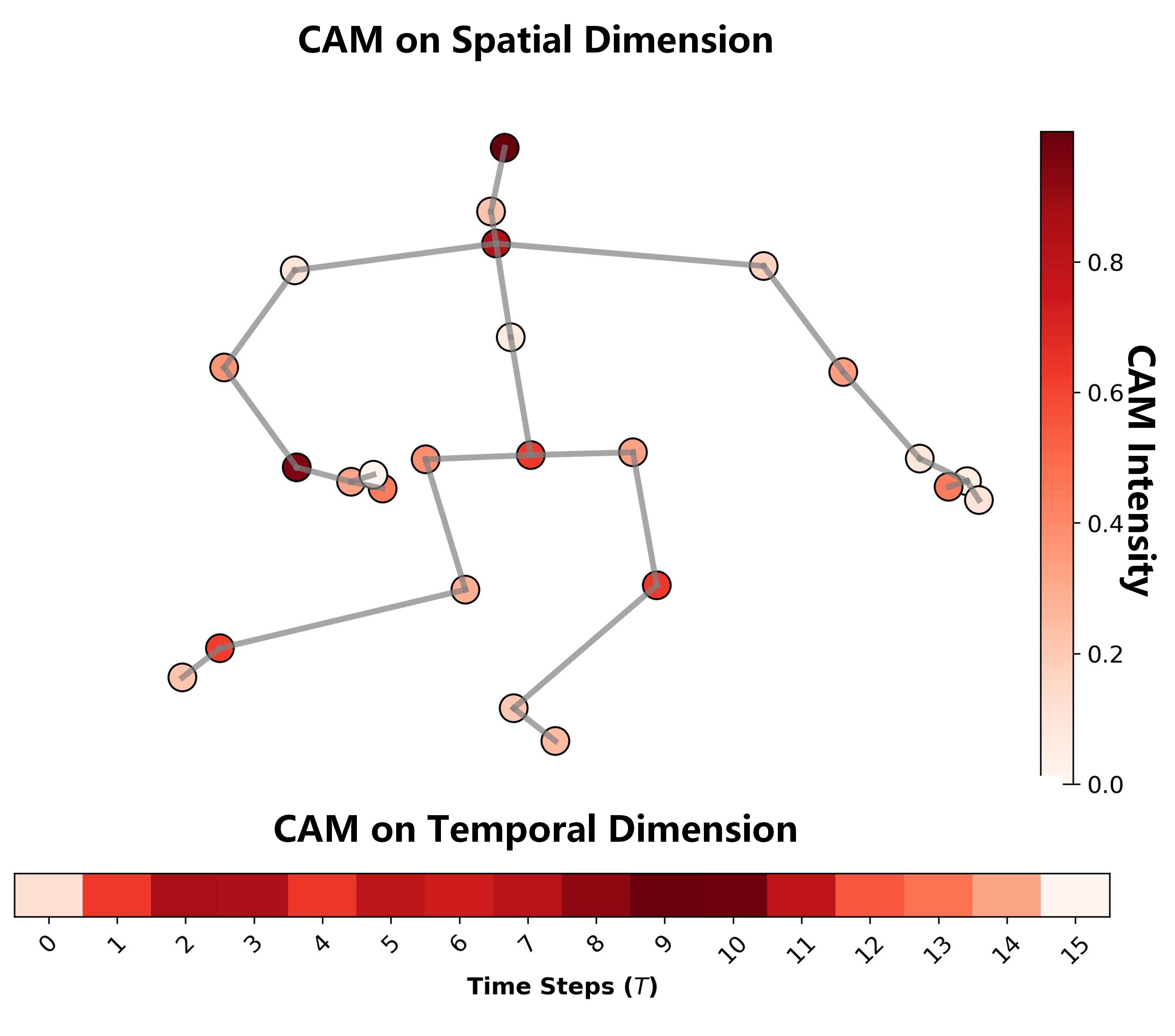}
    \label{fig:cam_25}
}
\caption{\textcolor{black}{\textbf{CAM visualizations on spatial and temporal dimensions.} 
Each subfigure corresponds to one action class, with the upper skeleton plot showing CAM intensities across spatial joints and the lower bar showing temporal contributions. Warmer colors (yellow/white) denote stronger activation responses, indicating which joints and time frames are most critical for the model’s classification.}}
\label{fig:cam_visualizations}
\end{figure}

\textit{Case A: Drop.}
For the Drop action, the CAM in Fig.\ref{fig:cam_4} shows that MK-SGN focuses on the upper-limb joints, particularly the shoulders, elbows, and wrists, which are crucial during the release phase. The trunk joints are moderately activated, reflecting their role in posture stabilization, while the lower-limb joints receive slight attention. In the temporal domain, peak activations occur around frames 7–8, when the object is released. 

\textit{Case B: Put on Glasses.}
For the Put on Glasses action (Fig.\ref{fig:cam_17}), the CAM highlights the face-related joints, especially the eyes, which receive the highest activation, aligning with their significance in the action. The lower limbs and trunk show little attention, reflecting their relatively low involvement. In the temporal domain, activations peak around frames 7–8, capturing the critical moment when the glasses are placed. 

\textit{Case C: Hopping.}
For the Hopping action (Fig.\ref{fig:cam_25}), MK-SGN concentrates attention on the lower limbs, particularly the knees and ankles, which dominate the motion. The upper body receives slight attention, consistent with the nature of the action. In the temporal domain, intermittent peaks are observed, especially near frames 9–10, reflecting the recurrent critical moments of the take-off process captured by the model.

In all three cases, MK-SGN's CAM visualizations confirm its ability to focus on the most relevant spatial and temporal features for each action. The model selectively attends to key joints and critical time steps, proving its effectiveness in capturing the necessary cues for accurate action recognition.

\paragraph{\textbf{t-SNE Visualization of Action Embeddings.}}

\begin{figure}[t]
\centering
\includegraphics[width=\linewidth]{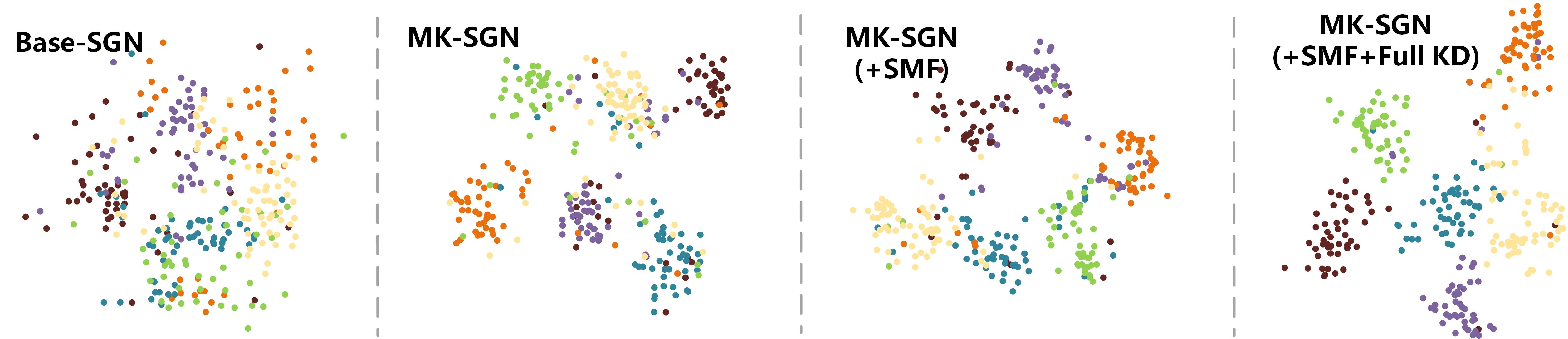}
\caption{\textcolor{black}{\textbf{t-SNE visualization of action embeddings.} The plot shows the embeddings of six action samples across four variants of MK-SGN: (a) Base-SGN, (b) MK-SGN, (c) MK-SGN (+SMF), and (d) MK-SGN (+SMF+Full KD). The clusters demonstrate the effectiveness of each model variant in capturing and distinguishing actions representation.}}
\label{fig:t-sne}
\end{figure}

Fig.~\ref{fig:t-sne} presents the t-SNE visualization of six sample actions across four MK-SGN variants: Base-SGN, MK-SGN, MK-SGN (+SMF), and MK-SGN (+SMF+Full KD).

For Base-SGN, the embeddings show scattered distributions with slight separation between actions, indicating limited feature discrimination. MK-SGN, on the other hand, demonstrates significant improvement, with clearer clusters forming around individual actions, highlighting the effectiveness of spiking graph convolution and attention mechanisms. The addition of SMF further refines the clusters, showing improved separation between actions. Finally, the Full KD variant achieves the most compact clusters, suggesting that knowledge distillation enhances the model’s ability to learn more discriminative action representations.

This analysis underscores the incremental improvements in action representation, validating the effectiveness of each component in enhancing the model's ability to capture and distinguish complex action features.
\begin{table*}[t]
\centering
\caption{Model’s architecture for NTU-RGB+D}
\setlength{\tabcolsep}{1.3pt} 
\footnotesize 
\begin{tabular}{c|c|c|c|c|c|c|c}
\hline
\textbf{Layer} & \textbf{GCN} & \textbf{GCN Structure} & \textbf{Base-SGN} & \textbf{MK-SGN} & \textbf{SGN Structure} & \textbf{Spikformer} & \textbf{Spikformer Structure} \\
 & \textbf{(Input/Output)} & \textbf{Layer Composition} & \textbf{(Input/Output)} & \textbf{(Input/Output)} & \textbf{Layer Composition} & \textbf{(Input/Output)} & \textbf{Layer Composition} \\
\hline
Fire layer & - & - & 3/3 & 3/3 & SC-(SMF) & 3/256 & SPS \\
1 & 3/64 & GC-TC & 3/64 & 3/64 & SGC-STC & 256/256 & SSA-MLP \\

2 & 64/64 & GC-TC & 64/64 & 64/64 & SGC-STC & 256/256 & SSA-MLP \\

3 & 64/64 & GC-TC & 64/64 & 64/128(stride=2) & SGC-STC & 256/256 & SSA-MLP \\

4 & 64/64& GC-TC & 64/64 & 128/128 & SGC-STC & 256/256 & SSA-MLP \\

5 & 64/128(stride=2) & GC-TC & 64/128(stride=2) & 128/256(stride=2) & SGC-STC & 256/256 & SSA-MLP \\

6 & 128/128 & GC-TC & 128/128 & 256/256 & SGC-STC & 256/256 & SSA-MLP \\

7 & 128/128 & GC-TC & 128/128 & - & SGC-STC & - & - \\

8 & 128/256(stride=2) & GC-TC & 128/256(stride=2) & - & SGC-STC & - & - \\

9 & 256/256 & GC-TC & 256/256 & - & SGC-STC & - & - \\

10 & 256/256 & GC-TC & 256/256 & - & SGC-STC & - & - \\
\hline
final & 256/class number & GAP-FC & 256/class number & 256/class number & SN-GAP-FC & 256/class number & SN-GAP-FC \\
\hline
\end{tabular}
\label{7}
\end{table*}
}
\subsection{Comparison with State-of-the-Art Methods}
This section comprehensively compares our proposed MK-SGN and SOTA methods for skeleton-based action recognition. The comparison is divided into two parts: (1).model architecture comparison and (2).metric comparison. 

In the model architecture comparison, we focus on the configuration and layer design of MK-SGN in contrast to vanilla GCN and SNN models, using the NTU-RGB+D 60 dataset as a case study. By analyzing the architectural differences, this comparison provides a foundation for the subsequent evaluation of performance metrics, ensuring a fair and consistent basis for comparison.

In the metric comparison, we evaluate MK-SGN against SOTA models across several dimensions, including accuracy, number of operations (OPs), and theoretical energy consumption. This evaluation encompasses publicly available results and those reproduced under consistent experimental settings, providing a comprehensive perspective on MK-SGN's ability to balance high performance with energy efficiency, particularly in energy-constrained environments.

By integrating both structural and performance analyses, we aim to demonstrate the superiority of MK-SGN in achieving competitive results with significantly reduced energy and computational overhead, establishing its practicality and effectiveness in real-world applications.

\subsubsection{Model Architecture Comparison}

Table~\ref{7} summarizes the architectural differences between our proposed MK-SGN and various models, including the unified framework of Vanilla GCN \cite{yan2018,shi2019,cheng2020,chen2021} and Spikformer \cite{zhou2022}, as applied to the NTU-RGB+D dataset. This comparison highlights how each model is structured to handle skeleton-based action recognition tasks.

The Vanilla GCN model is a classic ANN-based approach, utilizing GC layers to capture spatial dependencies and TC layers for temporal modeling. Its architecture typically follows a pyramid design with 10 layers of GC-TC operations, featuring progressively increasing feature dimensions and temporal downsampling through strides, culminating in a GAP-FC layer for final classification. 

The Base-SGN* model serves as the baseline SNN structure, leveraging SGC and STC layers to process spike-form data. It follows a similar configuration to GCN but is adapted for spiking computation.

The Spikformer model, a spiking-based Transformer, incorporates SSA layers and Multilayer Perceptron (MLP) components for feature extraction. For this study, we modified Spikformer to align with MK-SGN in terms of layer count and hidden dimensions, ensuring a fair comparison of metrics such as energy efficiency and accuracy. This adjustment enables us to evaluate the performance of Spikformer in a setting directly comparable to MK-SGN. Similarly, other network variations, such as Spiking-driven-former, Spiking-driven-former V2, and Spiking Wavelet Transformer, used in the study are adapted to maintain consistency across evaluations.

Finally, the MK-SGN  model stands out with its innovative spiking-specific modules, including the SSC and SMF modules, which enhance multimodal integration and energy efficiency. The SGC-STC layers are optimized for spike-form data, enabling MK-SGN to balance high accuracy with reduced computational and energy requirements, setting it apart as an advanced solution for real-world applications.

This architectural comparison provides the groundwork for the subsequent metric analysis, ensuring that the reproduced models are aligned in structure and settings to allow for fair and meaningful performance evaluations.
\subsubsection{Metric Comparison}
\begin{table*}[t]
\setlength{\tabcolsep}{3.5pt} 
\caption{Comparative results on NTU RGB+D and NTU RGB+D 120. We evaluate our model in terms of classification accuracy (\%). XSub, XView and XSet represent Cross-Subject, Cross-View and Cross-Setup splits.}
\footnotesize
\begin{threeparttable}
\centering
\begin{tabular}{ccccccccccc}
\toprule
\multirow{2}{*}{\textbf{Method}} & \multirow{2}{*}{\textbf{Architecture}} & \multirow{2}{*}{\textbf{Backbone}} & \multirow{2}{*}{\textbf{Spike Time Step}}& \multicolumn{3}{c}{\textbf{Performance}} & \multicolumn{4}{c}{\textbf{Accuracy (\%)}} \\
\cmidrule(lr){5-7} \cmidrule(lr){8-11}
                                 &                                        &                                   &                                 & \textbf{Param} & \textbf{OPs} & \textbf{Power} & \textbf{XSub} & \textbf{XView} & \textbf{XSub} & \textbf{XSet} \\
                                 &                                        &                                   &                                 & \textbf{(M)}   & \textbf{(G)} & \textbf{(mJ)}  & \textbf{} & \textbf{}  & \textbf{(120)}& \textbf{(120)} \\
\midrule
Lie Group\cite{vemulapalli2014}                        & ANN                                    & CNN                               & /                               & /              & /            & /              & 50.1          & 52.8           & /             & /              \\
HBRNN \cite{du2015}                           & ANN                                    & RNN                               & /                               & /              & /            & /              & 59.1          & 64.0           & /             & /              \\
Deep LSTM\cite{shahroudy2016}                        & ANN                                    & LSTM                              & /                               & /              & /            & /              & 60.7          & 67.3           & /             & /              \\
ST-GCN \cite{yan2018}                          & ANN                                    & GCN                               & /                         & 3.1           & 3.48        & 16.01         & 81.5          & 88.3           & 70.7          & 73.2           \\
2S-AGCN \cite{shi2019}                         & ANN                                    & GCN                               & /                         & 3.48           & 37.32        & 182.87         & 88.5          & 95.0           & 82.5          & 84.2           \\
Shift-GCN \cite{cheng2020}                       & ANN                                    & GCN                               & /                         & /              & /          & /            & 90.7          & 96.5           & 85.9          & 87.6           \\
MS-G3D \cite{liu2020}                          & ANN                                    & GCN                               & /                         & 3.19           & 48.88        & 239.51        & 91.5          & 96.2           & 86.9          & 88.4           \\

\textcolor{black}{CD-JBF-GCN\cite{tu2022}} & 
\textcolor{black}{ANN} & 
\textcolor{black}{GCN} & 
\textcolor{black}{/} & 
\textcolor{black}{1.46} & 
\textcolor{black}{/} & 
\textcolor{black}{/} & 
\textcolor{black}{89.0} & 
\textcolor{black}{95.7} & 
\textcolor{black}{/} & 
\textcolor{black}{/} \\
CTR-GCN\cite{chen2021}                          & ANN                                    & GCN                               & /                          & 1.46           & 7.88         & 36.25          & 92.4          & 96.8           & 88.9          & 90.6           \\
\textcolor{black}{Info-GCN \cite{chi2022}} & 
\textcolor{black}{ANN} & 
\textcolor{black}{GCN} & 
\textcolor{black}{/} & 
\textcolor{black}{1.6} & 
\textcolor{black}{7.36} & 
\textcolor{black}{33.86} & 
\textcolor{black}{92.3} & 
\textcolor{black}{96.7} & 
\textcolor{black}{89.2} & 
\textcolor{black}{90.7} \\
\textcolor{black}{FR-Head\cite{zhou2023}} & 
\textcolor{black}{ANN} & 
\textcolor{black}{GCN} & 
\textcolor{black}{/} & 
\textcolor{black}{2.0} & 
\textcolor{black}{/} & 
\textcolor{black}{/} & 
\textcolor{black}{92.8} & 
\textcolor{black}{96.8} & 
\textcolor{black}{89.5} & 
\textcolor{black}{90.9} \\

\textcolor{black}{BlockGCN\cite{zhou2024}} & 
\textcolor{black}{ANN} & 
\textcolor{black}{GCN} & 
\textcolor{black}{/} & 
\textcolor{black}{1.3} & 
\textcolor{black}{7.08} & 
\textcolor{black}{32.57} & 
\textcolor{black}{93.1} & 
\textcolor{black}{97.0} & 
\textcolor{black}{90.3} & 
\textcolor{black}{91.5} \\

\midrule
Base-SGN*                        & SNN                                    & SGN                               & 4                          & 2.07           & 0.60         & 0.536          & 64.2          & 71.3           & 45.2          & 47.3           \\ 
Spikformer\cite{zhou2022}\textsuperscript{ICLR 2023}                      & SNN                                    & Spiking Transformer               & 4                         & 4.78           & 1.69         & 2.17           & 73.9          & 80.1           & 61.7          & 63.7           \\ 
Spike-driven Transformer\cite{yao2024a} \textsuperscript{NIPS2023}       & SNN                                    & Spiking Transformer               & 4                         & 4.77           & 1.57         & 1.93           & 73.4          & 80.6           & 62.3          & 64.1           \\ 
Spike-driven Transformer V2\cite{yao2024b}\textsuperscript{ICLR 2024}    & SNN                                    & Spiking Transformer               & 4                         & 11.47          & 2.59         & 2.91           & 77.4          & 83.6           & 64.3          & 65.9           \\ 
Spiking Wavelet Transformer\cite{fang2025}\textsuperscript{ECCV 2025}    & SNN                                    & Spiking Transformer               & 4                         & 3.24           & 1.48         & 2.01           & 74.7          & 81.2           & 63.5          & 64.7           \\  
\textcolor{black}{STAtten\cite{lee2025}\textsuperscript{CVPR 2025}} & 
\textcolor{black}{SNN} & 
\textcolor{black}{Spiking Transformer} & 
\textcolor{black}{4} & 
\textcolor{black}{3.19} & 
\textcolor{black}{1.98} & 
\textcolor{black}{2.48} & 
\textcolor{black}{72.8} & 
\textcolor{black}{79.7} & 
\textcolor{black}{60.3} & 
\textcolor{black}{61.7} \\

\rowcolor[gray]{0.9}
\textbf{MK-SGN (Ours)}                      & SNN                                    & SGN                               & 4                & 2.17           & \textbf{0.67}& \textbf{0.596} & \textbf{78.5} & \textbf{85.6}  & \textbf{67.8} & \textbf{69.5}  \\
\bottomrule
\end{tabular}
\begin{tablenotes}
\item \textit{Note:} FLOPs are calculated using the \href{https://github.com/facebookresearch/fvcore}{FVCORE} library, which provides precise computation of floating-point operations.
\end{tablenotes}
\end{threeparttable}
\label{2}
\end{table*}

\begin{table*}[t]
\centering
\begin{threeparttable}
\caption{Comparative results on NW-UCLA. We evaluate our model in terms of classification accuracy (\%). }
\setlength{\tabcolsep}{5pt} 
\footnotesize 
\begin{tabular}{ccccccccc}
\hline
\textbf{Method}               & \textbf{Architecture} & \textbf{Backbone} & \textbf{Times} & \textbf{Param.} & \textbf{OPs}   & \textbf{Power} & \textbf{Acc}   \\ 
                              &                       &                   &                & \textbf{(M)}    & \textbf{(G)}   & \textbf{(mJ)}  & \textbf{(\%)}  \\ \hline
Lie Group\cite{vemulapalli2014}                       & ANN                   & CNN               & /              & -               & -              & -              & 74.2           \\
Actionlet ensemble \cite{wang2013}           & ANN                   & CNN               & /              & -               & -              & -              & 76.0           \\
HBRNN-L\cite{veeriah2015}                       & ANN                   & RNN               & /              & -               & -              & -              & 78.5           \\
Ensemble TS-LSTM \cite{song2017}             & ANN                   & LSTM              & /              & -               & -              & -              & 89.2           \\
Shift-GCN  \cite{cheng2020}                       & ANN                   & GCN               & /              & -               & 0.7            & 3.22           & 94.6           \\
CTR-GCN  \cite{chen2021}                       & ANN                   & GCN               & /              & 1.46            & 2.48           & 12.15          & 96.5           \\ 
\textcolor{black}{Info-GCN \cite{chi2022}} & 
\textcolor{black}{ANN} & 
\textcolor{black}{GCN} & 
\textcolor{black}{/} & 
\textcolor{black}{1.6} & 
\textcolor{black}{2.31} & 
\textcolor{black}{10.64} & 
\textcolor{black}{96.5} \\
\textcolor{black}{FR-Head\cite{zhou2023}} & 
\textcolor{black}{ANN} & 
\textcolor{black}{GCN} & 
\textcolor{black}{/} & 
\textcolor{black}{2.0} & 
\textcolor{black}{/} & 
\textcolor{black}{/} & 
\textcolor{black}{96.8} \\

\textcolor{black}{BlockGCN\cite{zhou2024}} & 
\textcolor{black}{ANN} & 
\textcolor{black}{GCN} & 
\textcolor{black}{/} & 
\textcolor{black}{1.3} & 
\textcolor{black}{1.96} & 
\textcolor{black}{9.05} & 
\textcolor{black}{96.9} \\
\hline

Base-SGN*                     & SNN                   & SGN               & 4              & 2.07            & 0.151          & 0.195          & 86.4           \\
Spikformer\cite{zhou2022}\textsuperscript{ICLR 2023}                       & SNN                   & Spiking Transformer & 4              & 4.78            & 0.513          & 0.673          & 85.4           \\
Spike-driven Transformer\cite{yao2024a}\textsuperscript{NIPS2023}      & SNN                   & Spiking Transformer & 4              & 4.77            & 0.501          & 0.643          & 83.4           \\
Spike-driven Transformer V2\cite{yao2024b}\textsuperscript{ICLR 2024}   & SNN                   & Spiking Transformer & 4              & 11.47           & 0.754          & 0.914          & 89.4           \\
Spiking Wavelet Transformer\cite{fang2025}\textsuperscript{ECCV 2025}   & SNN                   & Spiking Transformer & 4              & 3.24            & 0.451          & 0.697          & 86.9           \\
\textcolor{black}{STAtten\cite{lee2025}\textsuperscript{CVPR 2025}}  & 
\textcolor{black}{SNN} & 
\textcolor{black}{Spiking Transformer} & 
\textcolor{black}{4} & 
\textcolor{black}{3.19} & 
\textcolor{black}{0.451} & 
\textcolor{black}{0.697} & 
\textcolor{black}{86.9} \\
\rowcolor[gray]{0.9}
\textbf{MK-SGN (Ours)}        & SNN                   & SGN               & \textbf{4}     & \textbf{2.17}   & \textbf{0.165} & \textbf{0.207} & \textbf{92.3}  \\ \hline
\end{tabular}
\begin{tablenotes}
\footnotesize
\item \textit{Note:} FLOPs are calculated using the \href{https://github.com/facebookresearch/fvcore}{FVCORE} library, which provides precise computation of floating-point operations.
\end{tablenotes}
\label{table:nw_ucla_results}
\end{threeparttable}
\end{table*}
\textcolor{black}{
We compare our proposed MK-SGN with recent state-of-the-art (SOTA) ANNs and SNNs in skeleton-based action recognition, focusing on two key metrics: OPs and theoretical energy consumption. The results are summarized in Table~\ref{2} and Table~\ref{table:nw_ucla_results}.}

\textcolor{black}{To ensure a fair comparison, results for existing multi-modal GCN methods use the maximum number of modalities specified in their respective papers, including joint, bone, joint motion, and bone motion. The OPs and theoretical energy consumption for these multi-modal methods are calculated by multiplying the single-stream values by the number of modalities used. Additionally, we reproduce the results of recent SOTA SNNs using only the joint modality for a direct comparison with MK-SGN. These models are adapted to align with MK-SGN in terms of layers and hidden dimensions, ensuring consistency across experimental settings.}

\textcolor{black}{The comparison, shown in Table~\ref{2}, highlights MK-SGN's exceptional balance between accuracy and energy efficiency. MK-SGN significantly reduces theoretical energy consumption, requiring only 0.596~mJ per sample, representing a 96.28\% reduction compared to ST-GCN (16.01~mJ), 99.67\% reduction compared to 2S-AGCN (182.87~mJ), and 99.75\% reduction compared to MS-G3D (239.51~mJ). For lightweight models like Shift-GCN (46~mJ), CTR-GCN (36.25~mJ)  and BlockGCN (32.57~mJ), MK-SGN reduces energy consumption by 98.70\%, 98.36\% and 98.75\%, respectively, demonstrating its superior efficiency across ANN-based GCN models.}

\textcolor{black}{Beyond energy savings, MK-SGN outperforms recent SNN models in both accuracy and efficiency. For NTU-RGB+D, MK-SGN achieves 78.5\% (XSub) and 85.6\% (XView), surpassing competitive SNN models such as Spikformer (73.9\% XSub, 80.1\% XView) and Spike-driven Transformer V2 (77.4\% XSub, 83.6\% XView). Notably, MK-SGN achieves this with significantly reduced energy consumption—only 0.596~mJ per sample, compared to Spikformer’s 2.17~mJ and Spike-driven Transformer V2’s 2.91~mJ. Furthermore, MK-SGN demonstrates a highly efficient design, with only 2.17M parameters, far fewer than Spike-driven Transformer V2’s 11.47M. On the NTU-RGB+D 120 dataset, MK-SGN continues to outperform SNN models, achieving 67.8\% (XSub) and 69.5\% (XSet), while Spikformer and Spike-driven Transformer V2 achieve lower accuracies, 61.7\% (XSub) and 63.7\% (XSet), and 64.3\% (XSub) and 65.9\% (XSet), respectively. This illustrates MK-SGN's ability to maintain competitive performance and efficiency across different datasets.}

\textcolor{black}{On the NW-UCLA dataset, as shown in Table~\ref{table:nw_ucla_results}, MK-SGN achieves 92.3\% accuracy, surpassing other SOTA SNN models like Spikformer (85.4\%) and Spike-driven Transformer V2 (89.4\%), while maintaining significantly lower energy consumption (0.207~mJ per sample), compared to Spike-driven Transformer V2 (0.914~mJ). This confirms MK-SGN's remarkable ability to balance accuracy and energy efficiency, especially in scenarios that demand low computational cost.}

\begin{figure}[t]
    \centering
    \includegraphics[width=\linewidth]{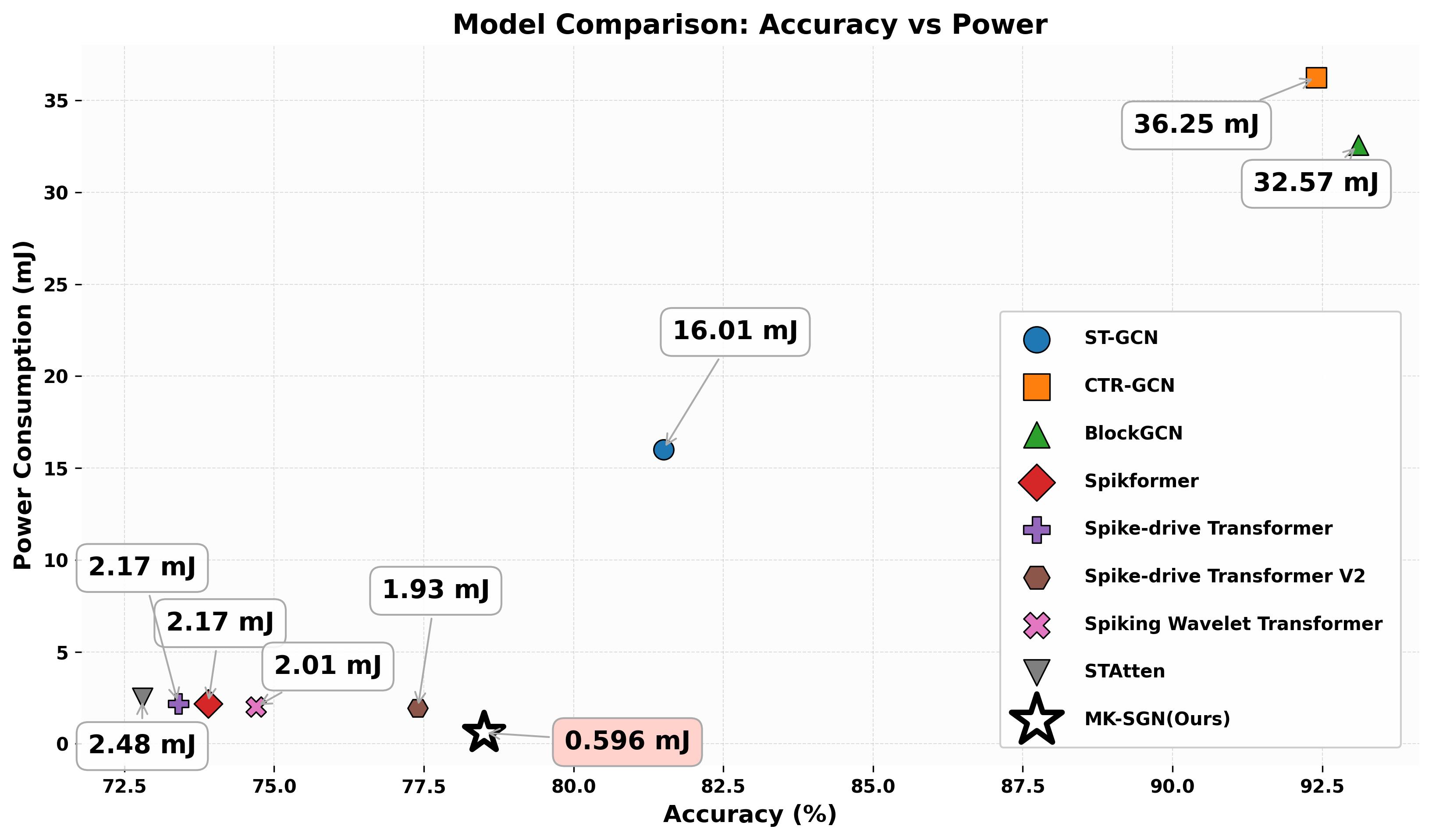}
    \caption{\textcolor{black}{Comparison of accuracy and power consumption for various models on the NTU-RGB+D CS dataset. MK-SGN outperforms other models in both accuracy and energy efficiency.}}
    \label{fig:accuracy_vs_power}
\end{figure}

\textcolor{black}{To further illustrate the synergy between accuracy and energy efficiency, we present the model comparison in Figure~\ref{fig:accuracy_vs_power}. The results, based on the NTU-RGB+D CS dataset, highlight MK-SGN’s superior performance in terms of both accuracy and power consumption.}

\textcolor{black}{These results validate MK-SGN as a reliable and energy-efficient solution for skeleton-based action recognition, excelling in scenarios requiring high accuracy and low computational overhead. The spiking-driven architecture of MK-SGN sets a new benchmark for balancing performance and efficiency in SNNs.}

{\color{black}
\subsection{Discussion}
\paragraph{\textbf{Principle Analysis: Accuracy and Efficiency Trade-off}}

MK-SGN achieves significant energy efficiency, but its accuracy is limited by the sparsity of SNNs. Unlike ANNs, which transmit information continuously, SNNs only transmit during spikes, leading to sparse information transfer and lower entropy.

Information entropy \( H(X) \) is defined as:

\begin{equation}
H(X) = - \sum_{i=1}^{n} p_i \log_2 p_i
\end{equation}

For ANNs, continuous activations result in higher entropy, enabling richer information flow. In contrast, SNNs transmit information only when neurons spike, which limits entropy and reduces the system's capacity to capture detailed patterns\cite{liu2025}. Despite its efficiency, MK-SGN’s accuracy is constrained by this sparsity. It outperforms other SNN models but still lags behind ANN-based models in high-precision tasks. The trade-off between energy efficiency and accuracy remains a challenge.

\paragraph{\textbf{Application Scenarios}}

While SNNs offer exceptional energy efficiency, their accuracy is often lower than that of GCNs, especially in tasks requiring detailed pattern recognition. SNNs are particularly suited for real-time applications where energy efficiency is a priority. For example, in wearable health monitoring systems, SNNs can be used for continuous gesture and activity recognition with relatively low energy consumption, making them ideal for long-term, battery-powered use. Similarly, in mobile robotics, SNNs can process sensor data efficiently, enabling real-time action recognition and decision-making on low-power devices, such as drones or robotic assistants. These applications require real-time, low-energy operation, where the trade-off between accuracy and energy efficiency is critical.

In contrast, GCNs excel in tasks requiring high accuracy and detailed spatial and temporal relationships, such as skeleton-based action recognition in complex environments. GCNs are ideal for scenarios where precise behavior analysis is necessary, such as in surveillance systems for human activity recognition in security or healthcare settings. For example, in smart home monitoring systems, GCNs can analyze detailed human motion patterns to detect falls, abnormal behaviors, or to improve user experience through activity-based control. Similarly, in sports analytics, GCNs can be used to analyze player movements in a game, providing insights into strategies, player health, and performance. These scenarios require high accuracy to capture the fine-grained details of human motion, which GCNs are well-suited for.

\paragraph{\textbf{Future Work}}

To enhance the performance of MK-SGN, future work should focus on optimizing spike coding, increasing spike rates, or exploring hybrid architectures with ANNs, like GCNs or CNNs, to enhance both accuracy and efficiency. SNNs can benefit from advanced spike encoding schemes that allow for richer information transmission while maintaining low energy consumption. Hybrid models that combine the strengths of SNNs and GCNs can offer a balance between real-time energy efficiency and high accuracy, especially in complex action recognition tasks.

Additionally, training methods tailored to SNNs should be developed to improve performance in high-accuracy tasks. Approaches such as spike-timing dependent plasticity (STDP) or supervised learning algorithms specifically designed for spiking neurons could lead to better convergence and higher recognition accuracy in the context of skeleton-based action recognition. These advancements will help address the current limitations and broaden the application of SNN-based models in real-world scenarios.

}

\section{Conclusion}
In this work, we propose the Spiking Graph Convolutional Network (MK-SGN), a novel architecture that combines the strengths of SNNs and GCNs for skeleton-based action recognition. To the best of our knowledge, this framework is the first to effectively integrate spiking neural computation with graph-based modeling, offering a new perspective on energy-efficient and high-performance recognition methods.

The MK-SGN addresses the critical challenge of high energy consumption in traditional GCN-based methods while maintaining competitive performance. Leveraging multimodal fusion and knowledge distillation, MK-SGN optimizes spiking neural computation and enhances the ability to extract and integrate meaningful features across modalities. These innovations enable MK-SGN to substantially decrease energy consumption across three benchmark datasets for skeleton action recognition, achieving an impressive reduction of over 98\% compared to the most effective ANN-based methods.

Moreover, MK-SGN consistently outperforms SOTA SNN models in recognition accuracy, achieving an average improvement of 5\%. This demonstrates its ability to balance energy consumption with recognition performance, making it well-suited for resource-constrained applications.{\color{black}However, the accuracy of MK-SGN is still limited by the sparsity inherent in SNNs. While the model achieves significant energy efficiency, its performance in high-precision tasks still lags behind traditional ANN-based methods, primarily due to the sparse information transfer characteristic of spiking networks.

Future work should focus on addressing these limitations by exploring advanced spike coding techniques to improve information transmission and enhance accuracy. Additionally, incorporating hybrid models that combine the strengths of SNNs and ANNs could help balance energy efficiency and accuracy more effectively. Optimizing spike rates and developing tailored training methods for SNNs, could also boost performance in high-accuracy scenarios.

Despite these challenges, MK-SGN offers a promising approach for energy-efficient skeleton-based action recognition, and its potential for real-time applications in resource-constrained environments positions it as a solid foundation for future exploration in spiking-based multimodal frameworks.}

\section{Acknowledgements} The National Natural Science Foundation of China NSFC-61976022 supports this work.

\end{document}